# A Tutorial on Dual Decomposition and Lagrangian Relaxation for Inference in Natural Language Processing

**Alexander M. Rush** SRUSH@CSAIL.MIT.EDU
*Computer Science and Artificial Intelligence Laboratory*
*Massachusetts Institute of Technology*
*Cambridge, MA 02139, USA*

**Michael Collins** MCOLLINS@CS.COLUMBIA.EDU
*Department of Computer Science*
*Columbia University*
*New York, NY 10027, USA*

## Abstract

Dual decomposition, and more generally Lagrangian relaxation, is a classical method for combinatorial optimization; it has recently been applied to several inference problems in natural language processing (NLP). This tutorial gives an overview of the technique. We describe example algorithms, describe formal guarantees for the method, and describe practical issues in implementing the algorithms. While our examples are predominantly drawn from the NLP literature, the material should be of general relevance to inference problems in machine learning. A central theme of this tutorial is that Lagrangian relaxation is naturally applied in conjunction with a broad class of combinatorial algorithms, allowing inference in models that go significantly beyond previous work on Lagrangian relaxation for inference in graphical models.

## 1. Introduction

In many problems in statistical natural language processing, the task is to map some input $x$ (e.g., a string) to some structured output $y$ (e.g., a parse tree). This mapping is often defined as

$$y^* = \operatorname*{argmax}_{y \in \mathcal{Y}} h(y) \quad (1)$$

where $\mathcal{Y}$ is a finite set of possible structures for the input $x$, and $h : \mathcal{Y} \rightarrow \mathbb{R}$ is a function that assigns a score $h(y)$ to each $y$ in $\mathcal{Y}$. For example, in part-of-speech tagging, $x$ would be a sentence, and $\mathcal{Y}$ would be the set of all possible tag sequences for $x$; in parsing, $x$ would be a sentence and $\mathcal{Y}$ would be the set of all parse trees for $x$; in machine translation, $x$ would be a source-language sentence and $\mathcal{Y}$ would be the set of all possible translations for $x$. The problem of finding $y^*$ is referred to as the decoding problem. The size of $\mathcal{Y}$ typically grows exponentially with respect to the size of the input $x$, making exhaustive search for $y^*$ intractable.

This paper gives an overview of decoding algorithms for NLP based on *dual decomposition*, and more generally, *Lagrangian relaxation*. Dual decomposition leverages the observation that many decoding problems can be decomposed into two or more sub-problems, together with linear constraints that enforce some notion of agreement between solutions to the different problems. The sub-problems are chosen such that they can be solved efficiently using exact combinatorial algorithms. The agreement constraints are incorporated using Lagrange multipliers, and an iterative algorithm—for example, a subgradient algorithm—is used to minimize the resulting dual. Dual decomposition algorithms have the following properties:

- They are typically simple and efficient. For example, subgradient algorithms involve two steps at each iteration: first, each of the sub-problems is solved using a combinatorial algorithm; second, simple additive updates are made to the Lagrange multipliers.
- They have well-understood formal properties, in particular through connections to linear programming (LP) relaxations.
- In cases where the underlying LP relaxation is tight, they produce an exact solution to the original decoding problem, with a certificate of optimality.[1] In cases where the underlying LP is not tight, heuristic methods can be used to derive a good solution; alternatively, constraints can be added incrementally until the relaxation is tight, at which point an exact solution is recovered.

Dual decomposition, where two or more combinatorial algorithms are used, is a special case of Lagrangian relaxation (LR). It will be useful to also consider LR methods that make use of a *single* combinatorial algorithm, together with a set of linear constraints that are again incorporated using Lagrange multipliers. The use of a single combinatorial algorithm is qualitatively different from dual decomposition approaches, although the techniques are very closely related.

Lagrangian relaxation has a long history in the combinatorial optimization literature, going back to the seminal work of Held and Karp (1971), who derive a relaxation algorithm for the traveling salesman problem. Initial work on Lagrangian relaxation/dual decomposition for decoding in statistical models focused on the MAP problem in Markov random fields (Komodakis, Paragios, & Tziritas, 2007, 2011). More recently, decoding algorithms have been derived for several models in statistical NLP, including models that combine a weighted context-free grammar (WCFG) with a finite-state tagger (Rush, Sontag, Collins, & Jaakkola, 2010); models that combine a lexicalized WCFG with a discriminative dependency parsing model (Rush et al., 2010); head-automata models for non-projective dependency parsing (Koo, Rush, Collins, Jaakkola, & Sontag, 2010); alignment models for statistical machine translation (DeNero & Macherey, 2011); models for event extraction (Riedel & McCallum, 2011); models for combined CCG parsing and supertagging (Auli & Lopez, 2011); phrase-based models for statistical machine translation (Chang & Collins, 2011); syntax-based models for statistical machine translation (Rush & Collins, 2011); models for semantic parsing (Das, Martins, & Smith, 2012); models for parsing and tagging that make use of document-level constraints (Rush, Reichart, Collins, & Globerson, 2012); models for the coordination problem in natural language parsing (Hanamoto, Matsuzaki, & Tsujii, 2012); and models based on the intersection of weighted automata (Paul & Eisner, 2012). We will give an overview of several of these algorithms in this paper.

While our focus is on examples from natural language processing, the material in this tutorial should be of general relevance to inference problems in machine learning. There is clear relevance to the problem of inference in graphical models, as described for example by Komodakis et al. (2007, 2011); however one central theme of this tutorial is that Lagrangian relaxation is naturally

1. A certificate of optimality is information that allows a proof of optimality of the solution to be constructed in polynomial time.

applied in conjunction with a much broader class of combinatorial algorithms than max-product belief propagation, allowing inference in models that go significantly beyond graphical models.

The remainder of this paper is structured as follows. Section 2 describes related work. Section 3 gives a formal introduction to Lagrangian relaxation. Section 4 describes a dual decomposition algorithm (from Rush et al., 2010) for decoding a model that combines a weighted context-free grammar with a finite-state tagger. This algorithm will be used as a running example throughout the paper. Section 5 describes formal properties of dual decomposition algorithms. Section 6 gives further examples of algorithms, and section 7 describes practical issues. Section 8 gives an overview of work on alternative optimization methods to the subgradient methods described in this tutorial. Finally, section 9 describes the relationship to LP relaxations, and describes tightening methods.

## 2. Related Work

This tutorial draws on ideas from the fields of combinatorial optimization, machine learning, and natural language processing. In this section, we give a summary of work from these fields that is relevant to the methods we will describe.

### 2.1 Combinatorial Optimization

Lagrangian relaxation (LR) is a widely used method in combinatorial optimization, going back to the seminal work of Held and Karp (1971) on the traveling salesman problem. See the work of Lemaréchal (2001) and Fisher (1981) for surveys of LR methods, and the textbook of Korte and Vygen (2008) for background on combinatorial optimization. Decomposing linear and integer linear programs is also a fundamental technique in the optimization community (Dantzig & Wolfe, 1960; Everett III, 1963). There is a very direct relationship between LR algorithms and linear programming relaxations of combinatorial optimization problems; again, see the textbook of Korte and Vygen.

### 2.2 Belief Propagation, and Linear Programming Relaxations for Inference in MRFs

There has been a large amount of research on the MAP inference problem in Markov random fields (MRFs). For tree-structured MRFs, max-product belief propagation (max-product BP) (Pearl, 1988) gives exact solutions. (Max-product BP is a form of dynamic programming, which is closely related to the Viterbi algorithm.) For general MRFs where the underlying graph may contain cycles, the MAP problem is NP-hard: this has led researchers to consider a number of approximate inference algorithms. Early work considered loopy variants of max-product BP (see for example Felzenszwalb & Huttenlocher, 2006, for the application of loopy max-product BP to problems in computer vision); however, these methods are heuristic, lacking formal guarantees.

More recent work has considered methods based on linear programming (LP) relaxations of the MAP problem. See the work of Yanover, Meltzer, and Weiss (2006), or section 1.6 of the work of Sontag, Globerson, and Jaakkola (2010), for a description. Methods based on LP relaxations have the benefit of stronger guarantees than loopy belief propagation. Inference is cast as an optimization problem, for example the problem of minimizing a dual. Since the dual problem is convex, convergence results from convex optimization and linear programming can be leveraged directly. One particularly appealing feature of these methods is that certificates of optimality can be given when the exact solution to the MAP problem is found.

Komodakis et al. (2007, 2011) describe a dual decomposition method that provably optimizes the dual of an LP relaxation of the MAP problem, using a subgradient method. This work is a crucial reference for this tutorial. (Note that in addition, Johnson, Malioutov, & Willsky, 2007, also describes LR methods for inference in MRFs.)

In this tutorial we focus on subgradient algorithms for optimization of the dual objective. See section 8 for a discussion of alternative optimization approaches that have been developed within the machine learning community.

### 2.3 Combinatorial Algorithms in Belief Propagation

A central idea in the algorithms we describe is the use of combinatorial algorithms other than max-product BP. This idea is closely related to earlier work on the use of combinatorial algorithms within belief propagation, either for the MAP inference problem (Duchi, Tarlow, Elidan, & Koller, 2007), or for computing marginals (Smith & Eisner, 2008). These methods generalize loopy BP in a way that allows the use of combinatorial algorithms. Again, we argue that methods based on Lagrangian relaxation are preferable to variants of loopy BP, as they have stronger formal guarantees.

### 2.4 Linear Programs for Decoding in Natural Language Processing

Dual decomposition and Lagrangian relaxation are closely related to integer linear programming (ILP) approaches, and linear programming relaxations of ILP problems. Several authors have used integer linear programming directly for solving challenging problems in NLP. Germann, Jahr, Knight, Marcu, and Yamada (2001) use ILP to test the search error of a greedy phrase-based translation system on short sentences. Roth and Yih (2005) formulate a constrained sequence labeling problem as an ILP and decode using a general-purpose solver. Lacoste-Julien, Taskar, Klein, and Jordan (2006) describe a quadratic assignment problem for bilingual word alignment and then decode using an ILP solver. Both the work of Riedel and Clarke (2006) and Martins, Smith, and Xing (2009) formulates higher-order non-projective dependency parsing as an ILP. Riedel and Clarke decode using an ILP method where constraints are added incrementally. Martins et al. solve an LP relaxation and project to a valid dependency parse. Like many of these works, the method presented in this tutorial begins with an ILP formulation of the decoding problem; however, instead of employing a general-purpose solver we aim to speed up decoding by using combinatorial algorithms that exploit the underlying structure of the problem.

## 3. Lagrangian Relaxation and Dual Decomposition

This section first gives a formal description of Lagrangian relaxation, and then gives a description of dual decomposition, an important special case of Lagrangian relaxation. The descriptions we give are deliberately concise. The material in this section is not essential to the remainder of this paper, and may be safely skipped by the reader, or returned to in a second reading. However the descriptions here may be useful for those who would like to immediately see a formal treatment of Lagrangian relaxation and dual decomposition. All of the algorithms in this paper are special cases of the framework described in this section.

### 3.1 Lagrangian Relaxation

We assume that we have some finite set $\mathcal{Y}$, which is a subset of $\mathbb{R}^d$. The score associated with any vector $y \in \mathcal{Y}$ is

$$h(y) = y \cdot \theta$$

where $\theta$ is also a vector in $\mathbb{R}^d$. The decoding problem is to find

$$y^* = \operatorname*{argmax}_{y \in \mathcal{Y}} h(y) = \operatorname*{argmax}_{y \in \mathcal{Y}} y \cdot \theta \tag{2}$$

Under these definitions, each structure $y$ is represented as a $d$-dimensional vector, and the score for a structure $y$ is a linear function, namely $y \cdot \theta$. In practice, in structured prediction problems $y$ is very often a binary vector (i.e., $y \in \{0, 1\}^d$) representing the set of parts[2] present in the structure $y$. The vector $\theta$ then assigns a score to each part, and the definition $h(y) = y \cdot \theta$ implies that the score for $y$ is a sum of scores for the parts it contains.

We will assume that the problem in Eq. 2 is computationally challenging. In some cases, it might be an NP-hard problem. In other cases, it might be solvable in polynomial time, but with an algorithm that is still too slow to be practical.

The first key step in Lagrangian relaxation will be to choose a finite set $\mathcal{Y}' \subset \mathbb{R}^d$ that has the following properties:

- $\mathcal{Y} \subset \mathcal{Y}'$. Hence $\mathcal{Y}'$ contains all vectors found in $\mathcal{Y}$, and in addition contains some vectors that are not in $\mathcal{Y}$.
- For any value of $\theta \in \mathbb{R}^d$, we can easily find

  $$\operatorname*{argmax}_{y \in \mathcal{Y}'} y \cdot \theta$$

  (Note that we have replaced $\mathcal{Y}$ in Eq. 2 with the larger set $\mathcal{Y}'$.) By "easily" we mean that this problem is significantly easier to solve than the problem in Eq. 2. For example, the problem in Eq. 2 might be NP-hard, while the new problem is solvable in polynomial time; or both problems might be solvable in polynomial time, but with the new problem having significantly lower complexity.
- Finally, we assume that

  $$\mathcal{Y} = \{y : y \in \mathcal{Y}' \text{ and } Ay = b\} \tag{3}$$

  for some $A \in \mathbb{R}^{p \times d}$ and $b \in \mathbb{R}^p$. The condition $Ay = b$ specifies $p$ linear constraints on $y$. We will assume that the number of constraints, $p$, is polynomial in the size of the input.

The implication here is that the linear constraints $Ay = b$ need to be added to the set $\mathcal{Y}'$, but these constraints considerably complicate the decoding problem. Instead of incorporating them as hard constraints, we will deal with these constraints using Lagrangian relaxation.

2. For example, in context-free parsing each part might correspond to a tuple $\langle A \to B\ C, i, k, j \rangle$ where $A \to B\ C$ is a context-free rule, and $i, k, j$ are integers specifying that non-terminal $A$ spans words $i \dots j$ in the input sentence, non-terminal $B$ spans words $i \dots k$, and non-terminal $C$ spans words $(k + 1) \dots j$. In finite-state tagging with a bigram tagging model each part might be a tuple $\langle A, B, i \rangle$ where $A, B$ are tags, and $i$ is an integer specifying that tag $B$ is seen at position $i$ in the sentence, and that tag $A$ is seen at position $(i - 1)$. See the work of Rush et al. (2010) for a detailed treatment of both of these examples.

We introduce a vector of Lagrange multipliers, $u \in \mathbb{R}^p$. The Lagrangian is

$$L(u, y) = y \cdot \theta + u \cdot (Ay - b)$$

This function combines the original objective function $y \cdot \theta$, with a second term that incorporates the linear constraints and the Lagrange multipliers. The dual objective is

$$L(u) = \max_{y \in \mathcal{Y}'} L(u, y)$$

and the dual problem is to find

$$\min_{u \in \mathbb{R}^p} L(u)$$

A common approach—which will be used in all algorithms in this paper—is to use a subgradient algorithm to minimize the dual. We set the initial Lagrange multiplier values to be $u^{(0)} = 0$. For $k = 1, 2, \ldots$ we then perform the following steps:

$$y^{(k)} = \operatorname*{argmax}_{y \in \mathcal{Y}'} L(u^{(k-1)}, y) \tag{4}$$

followed by

$$u^{(k)} = u^{(k-1)} - \delta_k (Ay^{(k)} - b) \tag{5}$$

where $\delta_k > 0$ is the step size at the $k$'th iteration. Thus at each iteration we first find a structure $y^{(k)}$, and then update the Lagrange multipliers, where the updates depend on $y^{(k)}$.

A crucial point is that $y^{(k)}$ can be found efficiently, because

$$\operatorname*{argmax}_{y \in \mathcal{Y}'} L(u^{(k-1)}, y) = \operatorname*{argmax}_{y \in \mathcal{Y}'} \left( y \cdot \theta + u^{(k-1)} \cdot (Ay - b) \right) = \operatorname*{argmax}_{y \in \mathcal{Y}'} y \cdot \theta'$$

where $\theta' = \theta + A^\top u^{(k-1)}$. Hence the Lagrange multiplier terms are easily incorporated into the objective function.

We can now state the following theorem:

**Theorem 1** *The following properties hold for Lagrangian relaxation:*

*a). For any $u \in \mathbb{R}^p$, $L(u) \geq \max_{y \in \mathcal{Y}} h(y)$.*

*b). Under a suitable choice of the step sizes $\delta_k$ (see section 5), $\lim_{k \to \infty} L(u^{(k)}) = \min_u L(u)$.*

*c). Define $y_u = \operatorname{argmax}_{y \in \mathcal{Y}'} L(u, y)$. If $\exists u$ such that $Ay_u = b$, then $y_u = \operatorname{argmax}_{y \in \mathcal{Y}} y \cdot \theta$ (i.e., $y_u$ is optimal).*

*In particular, in the subgradient algorithm described above, if for any $k$ we have $Ay^{(k)} = b$, then $y^{(k)} = \operatorname{argmax}_{y \in \mathcal{Y}} y \cdot \theta$.*

*d). $\min_u L(u) = \max_{\mu \in \mathcal{Q}} \mu \cdot \theta$, where the set $\mathcal{Q}$ is defined below.*

Thus part (a) of the theorem states that the dual value provides an upper bound on the score for the optimal solution, and part (b) states that the subgradient method successfully minimizes this upper bound. Part (c) states that if we ever reach a solution $y^{(k)}$ that satisfies the linear constraints, then we have solved the original optimization problem.

Part (d) of the theorem gives a direct connection between the Lagrangian relaxation method and an LP relaxation of the problem in Eq. 2. We now define the set $\mathcal{Q}$. First, define $\Delta$ to be the set of all distributions over the set $\mathcal{Y}'$:

$$\Delta = \{\alpha : \alpha \in \mathbb{R}^{|\mathcal{Y}'|}, \sum_{y \in \mathcal{Y}'} \alpha_y = 1, \forall y \; 0 \leq \alpha_y \leq 1\}$$

The convex hull of $\mathcal{Y}'$ is then defined as

$$\text{Conv}(\mathcal{Y}') = \{\mu \in \mathbb{R}^d : \exists \alpha \in \Delta \;\text{ s.t. }\; \mu = \sum_{y \in \mathcal{Y}'} \alpha_y y\}$$

Finally, define the set $\mathcal{Q}$ as follows:

$$\mathcal{Q} = \{y : y \in \text{Conv}(\mathcal{Y}') \;\text{ and }\; Ay = b\}$$

Note the similarity to Eq. 3: we have simply replaced $\mathcal{Y}'$ in Eq. 3 by the convex hull of $\mathcal{Y}'$. $\mathcal{Y}'$ is a subset of Conv($\mathcal{Y}'$), and hence $\mathcal{Y}$ is a subset of $\mathcal{Q}$. A consequence of the Minkowski-Weyl theorem (Korte & Vygen, 2008, Thm. 3.31) is that Conv($\mathcal{Y}'$) is a polytope (a bounded set that is specified by an intersection of a finite number of half spaces), and $\mathcal{Q}$ is therefore also a polytope. The problem

$$\max_{\mu \in \mathcal{Q}} \mu \cdot \theta$$

is therefore a linear program, and is a relaxation of our original problem, $\max_{y \in \mathcal{Y}} y \cdot \theta$.

Part (d) of theorem 1 is a direct consequence of duality in linear programming. It has the following implications:

- By minimizing the dual $L(u)$, we will recover the optimal value $\max_{\mu \in \mathcal{Q}} \mu \cdot \theta$ of the LP relaxation.
- If $\max_{\mu \in \mathcal{Q}} \mu \cdot \theta = \max_{y \in \mathcal{Y}} y \cdot \theta$ then we say that the LP relaxation is *tight*. In this case the subgradient algorithm is guaranteed[3] to find the solution to the original decoding problem,

$$y^* = \operatorname*{argmax}_{\mu \in \mathcal{Q}} \mu \cdot \theta = \operatorname*{argmax}_{y \in \mathcal{Y}} y \cdot \theta$$

- In cases where the LP relaxation is not tight, there are methods (e.g., see Nedić & Ozdaglar, 2009) that allow us to recover an approximate solution to the linear program, $\mu^* = \operatorname{argmax}_{\mu \in \mathcal{Q}} \mu \cdot \theta$. Alternatively, methods can be used to tighten the relaxation until an exact solution is obtained.

3. Under the assumption that there is unique solution $y^*$ to the problem $\max_{y \in \mathcal{Y}} y \cdot \theta$; if the solution is not unique then subtleties may arise.

### 3.2 Dual Decomposition

We now give a formal description of dual decomposition. As we will see, dual decomposition is a special case of Lagrangian relaxation;[4] however, it is important enough for the purposes of this tutorial to warrant its own description. Again, this section is deliberately concise, and may be safely skipped on a first reading.

We again assume that we have some finite set $\mathcal{Y} \subset \mathbb{R}^d$. Each vector $y \in \mathcal{Y}$ has an associated score

$$f(y) = y \cdot \theta^{(1)}$$

where $\theta^{(1)}$ is a vector in $\mathbb{R}^d$. In addition, we assume a second finite set $\mathcal{Z} \subset \mathbb{R}^{d'}$, with each vector $z \in \mathcal{Z}$ having an associated score

$$g(z) = z \cdot \theta^{(2)}$$

The decoding problem is then to find

$$\operatorname*{argmax}_{y \in \mathcal{Y}, z \in \mathcal{Z}} y \cdot \theta^{(1)} + z \cdot \theta^{(2)}$$

such that

$$Ay + Cz = b$$

where $A \in \mathbb{R}^{p \times d}$, $C \in \mathbb{R}^{p \times d'}$, $b \in \mathbb{R}^p$.

Thus the decoding problem is to find the optimal *pair* of structures, under the linear constraints specified by $Ay + Cz = b$. In practice, the linear constraints often specify agreement constraints between $y$ and $z$: that is, they specify that the two vectors are in some sense coherent.

For convenience, and to make the connection to Lagrangian relaxation clear, we will define the following sets:

$$\begin{aligned} \mathcal{W} &= \{(y, z) : y \in \mathcal{Y}, z \in \mathcal{Z}, Ay + Cz = b\} \\ \mathcal{W}' &= \{(y, z) : y \in \mathcal{Y}, z \in \mathcal{Z}\} \end{aligned}$$

It follows that our decoding problem is to find

$$\operatorname*{argmax}_{(y,z) \in \mathcal{W}} \left( y \cdot \theta^{(1)} + z \cdot \theta^{(2)} \right) \tag{6}$$

Next, we make the following assumptions:

4. Strictly speaking, Lagrangian relaxation can also be viewed as a special case of dual decomposition: in the formulation of this section we can set $\mathcal{Z} = \mathcal{Y}$, $\theta^{(2)} = 0$, and $C_{i,j} = 0$ for all $i, j$, thus recovering the Lagrangian relaxation problem from the previous section. In this sense Lagrangian relaxation and dual decomposition are equivalent (we can transform any Lagrangian relaxation problem to a dual decomposition problem, and vice versa). However, in our view dual decomposition is more naturally viewed as a special case of Lagrangian relaxation, in particular because the methods described in this tutorial go back to the work of Held and Karp (1971) (see section 6.3), which makes use of a single combinatorial algorithm. In addition, Lagrangian relaxation appears to be a more standard term in the combinatorial optimization literature: for example the textbook of Korte and Vygen (2008) has a description of Lagrangian relaxation but no mention of dual decomposition; there are several tutorials on Lagrangian relaxation in the combinatorial optimization literature (e.g., see Lemaréchal, 2001; Fisher, 1981), but we have found it more difficult to find direct treatments of dual decompositon. Note however that recent work in the machine learning and computer vision communities has often used the term dual decomposition (e.g., Sontag et al., 2010; Komodakis et al., 2007, 2011).

- For any value of $\theta^{(1)} \in \mathbb{R}^d$, we can easily find $\text{argmax}_{y \in \mathcal{Y}}\, y \cdot \theta^{(1)}$. Furthermore, for any value of $\theta^{(2)} \in \mathbb{R}^{d'}$, we can easily find $\text{argmax}_{z \in \mathcal{Z}}\, z \cdot \theta^{(2)}$. It follows that for any $\theta^{(1)} \in \mathbb{R}^d$, $\theta^{(2)} \in \mathbb{R}^{d'}$, we can easily find

$$(y^*, z^*) = \underset{(y,z) \in \mathcal{W}'}{\text{argmax}}\, y \cdot \theta^{(1)} + z \cdot \theta^{(2)} \tag{7}$$

by setting

$$y^* = \underset{y \in \mathcal{Y}}{\text{argmax}}\, y \cdot \theta^{(1)}, \quad z^* = \underset{z \in \mathcal{Z}}{\text{argmax}}\, z \cdot \theta^{(2)}$$

Note that Eq. 7 is closely related to the problem in Eq. 6, but with $\mathcal{W}$ replaced by $\mathcal{W}'$ (i.e., the linear constraints $Ay + Cz = b$ have been dropped). By "easily" we again mean that these optimization problems are significantly easier to solve than our original problem in Eq. 6.

It should now be clear that the problem is a special case of the Lagrangian relaxation setting, as described in the previous section. Our goal involves optimization of a linear objective, over the finite set $\mathcal{W}$, as given in Eq. 6; we can efficiently find the optimal value over a set $\mathcal{W}'$ such that $\mathcal{W}$ is a subset of $\mathcal{W}'$, and $\mathcal{W}'$ has dropped the linear constraints $Ay + Cz = b$.

The dual decomposition algorithm is then derived in a similar way to before. We introduce a vector of Lagrange multipliers, $u \in \mathbb{R}^p$. The Lagrangian is now

$$L(u, y, z) = y \cdot \theta^{(1)} + z \cdot \theta^{(2)} + u \cdot (Ay + Cz - b)$$

and the dual objective is

$$L(u) = \max_{(y,z) \in \mathcal{W}'} L(u, y, z)$$

A subgradient algorithm can again be used to find $\min_{u \in \mathbb{R}^p} L(u)$. We initialize the Lagrange multipliers to $u^{(0)} = 0$. For $k = 1, 2, \ldots$ we perform the following steps:

$$(y^{(k)}, z^{(k)}) = \underset{(y,z) \in \mathcal{W}'}{\text{argmax}}\, L(u^{(k-1)}, y, z)$$

followed by

$$u^{(k)} = u^{(k-1)} - \delta_k (Ay^{(k)} + Cz^{(k)} - b)$$

where each $\delta_k > 0$ is a stepsize.

Note that the solutions $y^{(k)}, z^{(k)}$ at each iteration are found easily, because it is easily verified that

$$\underset{(y,z) \in \mathcal{W}'}{\text{argmax}}\, L(u^{(k-1)}, y, z) = \left( \underset{y \in \mathcal{Y}}{\text{argmax}}\, y \cdot \theta'^{(1)}, \underset{z \in \mathcal{Z}}{\text{argmax}}\, z \cdot \theta'^{(2)}, \right)$$

where $\theta'^{(1)} = \theta^{(1)} + A^\top u^{(k-1)}$ and $\theta'^{(2)} = \theta^{(2)} + C^\top u^{(k-1)}$. Thus the dual decomposes into two easily solved maximization problems.

The formal properties for dual decomposition are very similar to those stated in theorem 1. In particular, it can be shown that

$$\min_{u \in \mathbb{R}^p} L(u) = \max_{(\mu,\nu) \in \mathcal{Q}} \mu \cdot \theta^{(1)} + \nu \cdot \theta^{(2)}$$

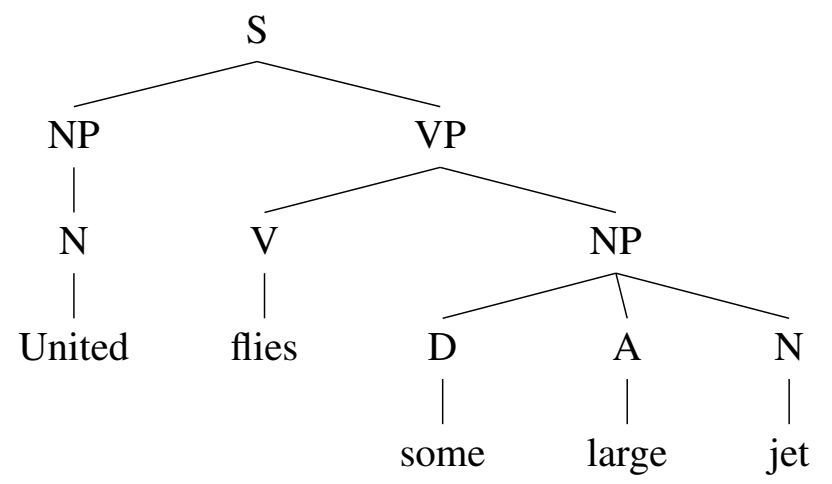

Figure 1: An example parse tree.

where the set $\mathcal{Q}$ is defined as

$$\mathcal{Q} = \{(\mu, \nu) : (\mu, \nu) \in \text{Conv}(\mathcal{W}') \;\text{ and }\; A\mu + C\nu = d\}$$

The problem

$$\max_{(\mu,\nu)\in\mathcal{Q}} \mu \cdot \theta^{(1)} + \nu \cdot \theta^{(2)}$$

is again a linear programming problem, and $L(u)$ is the dual of this linear program.

The descriptions of Lagrangian relaxation and dual decomposition that we have given are at a sufficient level of generality to include a very broad class of algorithms, including all those introduced in this paper. The remainder of this paper describes specific algorithms developed within this framework, describes experimental results and practical issues that arise, and elaborates more on the theory underlying these algorithms.

Note that in this section we have described the dual-decomposition approach with two components. The generalization to more than two components is relatively straightforward; for example see the work of Komodakis et al. (2007, 2011), see also the work of Martins, Smith, Figueiredo, and Aguiar (2011).

## 4. An Example: Integration of a Parser and a Finite-State Tagger

We next describe a dual decomposition algorithm for decoding under a model that combines a weighted context-free grammar and a finite-state tagger. The classical approach for this problem is to use a dynamic programming algorithm, based on the construction of Bar-Hillel, Perles, and Shamir (1964) for the intersection of a context-free language and a finite-state language. The dual decomposition algorithm has advantages over exhaustive dynamic programming, in terms of both efficiency and simplicity. We will use this dual decomposition algorithm as a running example throughout this tutorial.

We first give a formal definition of the problem, describe motivation for the problem, and describe the classical dynamic programming approach. We then describe the dual decomposition algorithm.

### 4.1 Definition of the Problem

Consider the problem of mapping an input sentence $x$ to a parse tree $y$. Define $\mathcal{Y}$ to be the set of all parse trees for $x$. The parsing problem is to find

$$y^* = \operatorname*{argmax}_{y \in \mathcal{Y}} h(y) \tag{8}$$

where $h(y)$ is the score for any parse tree $y \in \mathcal{Y}$.

We consider the case where $h(y)$ is the sum of two model scores: first, the score for $y$ under a weighted context-free grammar; and second, the score for the part-of-speech (POS) sequence in $y$ under a finite-state part-of-speech tagging model. More formally, we define $h(y)$ to be

$$h(y) = f(y) + g(l(y)) \tag{9}$$

where the functions $f$, $g$, and $l$ are defined as follows:

1. $f(y)$ is the score for $y$ under a weighted context-free grammar (WCFG). A WCFG consists of a context-free grammar with a set of rules $G$, and a scoring function $\theta : G \rightarrow \mathbb{R}$ that assigns a real-valued score to each rule in $G$. The score for an entire parse tree is the sum of scores for the rules it contains. As an example, consider the parse tree shown in Figure 1; for this tree,

   $$\begin{aligned} f(y) \;\; = \;\; & \theta(\text{S} \rightarrow \text{NP VP}) + \theta(\text{NP} \rightarrow \text{N}) + \theta(\text{N} \rightarrow \text{United}) \\ & + \theta(\text{VP} \rightarrow \text{V NP}) + \ldots \end{aligned}$$

   We remain agnostic as to how the scores for individual context-free rules are defined. As one example, in a probabilistic context-free grammar, we would define $\theta(\alpha \rightarrow \beta) = \log p(\alpha \rightarrow \beta|\alpha)$. As a second example, in a conditional random field (CRF) (Lafferty, McCallum, & Pereira, 2001) we would define $\theta(\alpha \rightarrow \beta) = w \cdot \phi(\alpha \rightarrow \beta)$ where $w \in \mathbb{R}^q$ is a parameter vector, and $\phi(\alpha \rightarrow \beta) \in \mathbb{R}^q$ is a feature vector representing the rule $\alpha \rightarrow \beta$.

2. $l(y)$ is a function that maps a parse tree $y$ to the sequence of part-of-speech tags in $y$. For the parse tree in Figure 1, $l(y)$ would be the sequence `N V D A N`.

3. $g(z)$ is the score for the part-of-speech tag sequence $z$ under an $m$'th-order finite-state tagging model. Under this model, if $z_i$ for $i = 1 \ldots n$ is the $i$'th tag in $z$, then

   $$g(z) = \sum_{i=1}^{n} \theta(i, z_{i-m}, z_{i-m+1}, \ldots, z_i)$$

   where $\theta(i, z_{i-m}, z_{i-m+1}, \ldots, z_i)$ is the score for the sub-sequence of tags $z_{i-m}, z_{i-m+1}, \ldots, z_i$ ending at position $i$ in the sentence.[5]

   We again remain agnostic as to how these $\theta$ terms are defined. As one example, $g(z)$ might be the log-probability for $z$ under a hidden Markov model, in which case

   $$\theta(i, z_{i-m} \ldots z_i) = \log p(z_i|z_{i-m} \ldots z_{i-1}) + \log p(x_i|z_i)$$

5. We define $z_i$ for $i \leq 0$ to be a special "start" POS symbol.

where $x_i$ is the $i$'th word in the input sentence. As another example, under a CRF we would have

$$\theta(i, z_{i-m} \ldots z_i) = w \cdot \phi(x, i, z_{i-m} \ldots z_i)$$

where $w \in \mathbb{R}^q$ is a parameter vector, and $\phi(x, i, z_{i-m} \ldots z_i)$ is a feature-vector representation of the sub-sequence of tags $z_{i-m} \ldots z_i$ ending at position $i$ in the sentence $x$.

The motivation for this problem is as follows. The scoring function $h(y) = f(y) + g(l(y))$ combines information from both the parsing model and the tagging model. The two models capture fundamentally different types of information: in particular, the part-of-speech tagger captures information about adjacent POS tags that will be missing under $f(y)$. This information may improve both parsing and tagging performance, in comparison to using $f(y)$ alone.[6]

Under this definition of $h(y)$, the conventional approach to finding $y^*$ in Eq. 8 is to construct a new context-free grammar that introduces sensitivity to surface bigrams (Bar-Hillel et al., 1964). Roughly speaking, in this approach (assuming a first-order tagging model) rules such as

$$\texttt{VP} \rightarrow \texttt{V} \quad \texttt{NP}$$

are replaced with rules such as

$$\texttt{VP}_{\texttt{N,N}} \rightarrow \texttt{V}_{\texttt{N,V}} \quad \texttt{NP}_{\texttt{V,N}} \tag{10}$$

where each non-terminal (e.g., `NP`) is replaced with a non-terminal that tracks the preceding and last POS tag relative to that non-terminal. For example, $\texttt{NP}_{\texttt{V,N}}$ represents a `NP` that dominates a sub-tree whose preceding POS tag was `V`, and whose last POS tag is `N`. The weights on the new rules are just context-free weights from $f(y)$. Furthermore, rules such as

$$\texttt{V} \rightarrow \texttt{flies}$$

are replaced with rules such as

$$\texttt{V}_{\texttt{N,V}} \rightarrow \texttt{flies}$$

The weights on these rules are the context-free weights from $f(y)$ plus the bigram tag weights from $g(z)$, in this example for the bigram `N V`. A dynamic programming parsing algorithm—for example the CKY algorithm—can then be used to find the highest scoring structure under the new grammar.

This approach is guaranteed to give an exact solution to the problem in Eq. 8; however it is often very inefficient. We have greatly increased the size of the grammar by introducing the refined non-terminals, and this leads to significantly slower parsing performance. As one example, consider the case where the underlying grammar is a CFG in Chomsky-normal form, with $G$ non-terminals, and where we use a 2nd order (trigram) tagging model, with $T$ possible part-of-speech tags. Define $n$ to be the length of the input sentence. Parsing with the grammar alone would take $O(G^3 n^3)$ time, for example using the CKY algorithm. In contrast, the construction of Bar-Hillel et al. (1964)

6. We have assumed that it is sensible, in a theoretical and/or empirical sense, to take a sum of the scores $f(y)$ and $g(l(y))$. This might be the case, for example, if $f(y)$ and $g(z)$ are defined through structured prediction models (e.g., conditional random fields), and their parameters are estimated jointly using discriminative methods. If $f(y)$ and $g(z)$ are log probabilities under a PCFG and HMM respectively, then from a strict probabilistic sense it does not make sense to combine their scores in this way: however in practice this may work well; for example, this type of log-linear combination of probabilistic models is widely used in approaches for statistical machine translation.

results in an algorithm with a run time of $O(G^3T^6n^3)$.[7] The addition of the tagging model leads to a multiplicative factor of $T^6$ in the runtime of the parser, which is a very significant decrease in efficiency (it is not uncommon for $T$ to take values of say 5 or 50, giving values for $T^6$ larger than $15,000$ or $15$ million). In contrast, the dual decomposition algorithm which we describe next takes $O(k(G^3n^3 + T^3n))$ time for this problem, where $k$ is the number of iterations required for convergence; in experiments, $k$ is often a small number. This is a very significant improvement in runtime over the Bar-Hillel et al. method.

### 4.2 The Dual Decomposition Algorithm

We now introduce an alternative formulation of the problem in Eq. 8, which will lead directly to the dual decomposition algorithm. Define $\mathcal{T}$ to be the set of all POS tags. Assume the input sentence has $n$ words. For any parse tree $y$, for any position $i \in \{1 \dots n\}$, for any tag $t \in \mathcal{T}$, we define $y(i,t) = 1$ if parse tree $y$ has tag $t$ at position $i$, $y(i,t) = 0$ otherwise. Similarly, for any tag sequence $z$, we define $z(i,t) = 1$ if the tag sequence has tag $t$ at position $i$, $0$ otherwise. As an example, the following parse tree and tag sequence have $y(4, A) = 1$ and $z(4, A) = 1$:

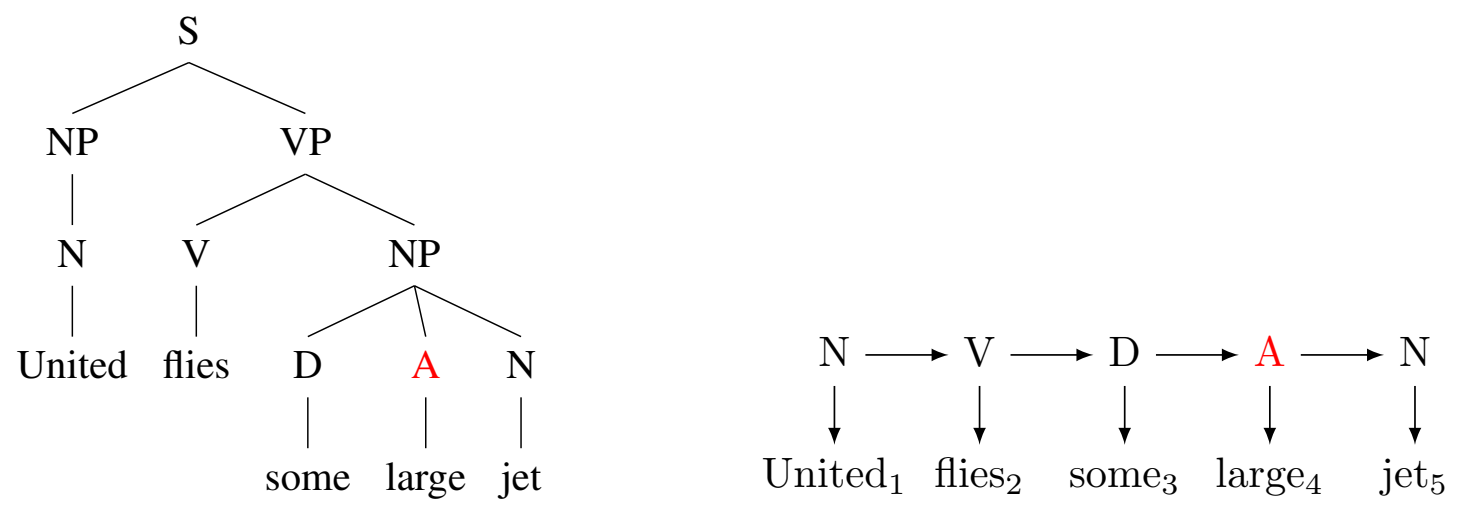

In addition, define $\mathcal{Z}$ to be the set of all possible POS tag sequences for the input sentence. We then introduce the following optimization problem:

**Optimization Problem 1** *Find*

$$\underset{y \in \mathcal{Y}, z \in \mathcal{Z}}{\operatorname{argmax}} f(y) + g(z) \tag{11}$$

*such that for all* $i \in \{1 \dots n\}$*, for all* $t \in \mathcal{T}$*,* $y(i,t) = z(i,t)$*.*

Thus we now find the best *pair* of structures $y$ and $z$ such that they share the same POS sequence. We define $(y^*, z^*)$ to be the pair of structures that achieve the argmax in this problem. The crucial

7. To be precise, assume we have a finite-state automaton with $Q$ states and a context-free chart with rule productions $\langle A \rightarrow B\ C, i, k, j\rangle$ for all $A, B, C \in \mathcal{G}$ and $1 \leq i < k < j \leq n$ as well as productions $\langle A \rightarrow w_i, i\rangle$ for all $A \in \mathcal{G}$ and $i \in \{1 \dots n\}$. (Here we use $w_i$ to refer to the $i$'th word in the sentence, and the set $\mathcal{G}$ to refer to the set of non-terminals in the grammar. It follows that $G = |\mathcal{G}|$.) Applying the Bar-Hillel intersection gives new rule productions $\langle A_{s_1,s_3} \rightarrow B_{s_1,s_2}\ C_{s_2,s_3}, i, k, j\rangle$ for $s_1, s_2, s_3 \in \{1 \dots Q\}$ as well as $\langle A_{s,t} \rightarrow w_i, i\rangle$ for $s, t \in \{1 \dots Q\}$ where $(s,t)$ is a valid state transition in the FSA. After intersection, we can count the free variables to see that there are $O(G^3n^3Q^3)$ rule productions, which implies that the CKY algorithm can find the best parse in $O(G^3n^3Q^3)$ time. In the case of tagging, a 2nd-order tagging model can be represented as an FSA with $|T|^2$ states, where each state represents the previous two tags. After intersection, this yields an $O(G^3n^3|T|^6)$ time algorithm.

claim, which is easily verified, is that $y^*$ is also the $\mathrm{argmax}$ to the problem in Eq. 8. In this sense, solving the new problem immediately leads to a solution to our original problem.

We then make the following two assumptions. Whether these assumptions are satisfied will depend on the definitions of $\mathcal{Y}$ and $f(y)$ (for assumption 1) and on the definitions of $\mathcal{Z}$ and $g(z)$ (for assumption 2). The assumptions hold when $f(y)$ is a WCFG and $g(z)$ is a finite-state tagger, but more generally they may hold for other parsing and tagging models.

**Assumption 1** *Assume that we introduce variables $u(i,t) \in \mathbb{R}$ for $i \in \{1 \ldots n\}$, and $t \in \mathcal{T}$. We assume that for any value of these variables, we can find*

$$\underset{y \in \mathcal{Y}}{\mathrm{argmax}} \left( f(y) + \sum_{i,t} u(i,t) y(i,t) \right)$$

*efficiently.*

*An example.* Consider a WCFG where the grammar is in Chomsky normal form. The scoring function is defined as

$$f(y) = \sum_{X \to YZ} c(y, X \to YZ)\theta(X \to YZ) + \sum_{i,t} y(i,t)\theta(t \to w_i)$$

where we write $c(y, X \to YZ)$ to denote the number of times that rule $X \to YZ$ is seen in the parse tree $y$, and as before $y(i,t) = 1$ if word $i$ has POS $t$, 0 otherwise (note that $y(i,t) = 1$ implies that the rule $t \to w_i$ is used in the parse tree). The highest scoring parse tree under $f(y)$ can be found efficiently, for example using the CKY parsing algorithm. We then have

$$\underset{y \in \mathcal{Y}}{\mathrm{argmax}} \left( f(y) + \sum_{i,t} u(i,t) y(i,t) \right) =$$
$$\underset{y \in \mathcal{Y}}{\mathrm{argmax}} \left( \sum_{X \to YZ} c(y, X \to YZ)\theta(X \to YZ) + \sum_{i,t} y(i,t)(\theta(t \to w_i) + u(i,t)) \right)$$

This $\mathrm{argmax}$ can again be found easily using the CKY algorithm, where the scores $\theta(t \to w_i)$ are simply replaced by new scores defined as $\theta'(t \to w_i) = \theta(t \to w_i) + u(i,t)$. $\square$

**Assumption 2** *Assume that we introduce variables $u(i,t) \in \mathbb{R}$ for $i \in \{1 \ldots n\}$, and $t \in \mathcal{T}$. We assume that for any value of these variables, we can find*

$$\underset{z \in \mathcal{Z}}{\mathrm{argmax}} \left( g(z) - \sum_{i,t} u(i,t) z(i,t) \right)$$

*efficiently.*

*An example.* Consider a 1st-order tagging model,

$$g(z) = \sum_{i=1}^{n} \theta(i, z_{i-1}, z_i)$$

**Initialization:** Set $u^{(0)}(i,t) = 0$ for all $i \in \{1 \ldots n\}$, $t \in \mathcal{T}$

**For** $k = 1$ **to** $K$

$y^{(k)} \leftarrow \text{argmax}_{y \in \mathcal{Y}} \left(f(y) + \sum_{i,t} u^{(k-1)}(i,t) y(i,t)\right)$ [Parsing]

$z^{(k)} \leftarrow \text{argmax}_{z \in \mathcal{Z}} \left(g(z) - \sum_{i,t} u^{(k-1)}(i,t) z(i,t)\right)$ [Tagging]

**If** $y^{(k)}(i,t) = z^{(k)}(i,t)$ for all $i, t$ **Return** $(y^{(k)}, z^{(k)})$

**Else** $u^{(k+1)}(i,t) \leftarrow u^{(k)}(i,t) - \delta_k (y^{(k)}(i,t) - z^{(k)}(i,t))$

Figure 2: The dual decomposition algorithm for integrated parsing and tagging. $\delta_k$ for $k = 1 \ldots K$ is the step size at the $k$'th iteration.

Then

$$\begin{aligned} & \underset{z \in \mathcal{Z}}{\text{argmax}} \left( g(z) - \sum_{i,t} u(i,t) z(i,t) \right) \\ = \quad & \underset{z \in \mathcal{Z}}{\text{argmax}} \left( \sum_{i=1}^{n} \theta(i, z_{i-1}, z_i) - \sum_{i,t} u(i,t) z(i,t) \right) \\ = \quad & \underset{z \in \mathcal{Z}}{\text{argmax}} \sum_{i=1}^{n} \theta'(i, z_{i-1}, z_i) \end{aligned}$$

where

$$\theta'(i, z_{i-1}, z_i) = \theta(i, z_{i-1}, z_i) - u(i, z_i)$$

This $\text{argmax}$ can be found efficiently using the Viterbi algorithm, where we have new $\theta'$ terms that incorporate the $u(i,t)$ values. □

Given these assumptions, the dual decomposition algorithm is shown in Figure 2. The algorithm manipulates a vector of variables $u = \{u(i,t) : i \in \{1 \ldots n\}, t \in \mathcal{T}\}$. We will soon see that each variable $u(i,t)$ is a Lagrange multiplier enforcing the constraint $y(i,t) = z(i,t)$ in our optimization problem. At each iteration the algorithm finds hypotheses $y^{(k)}$ and $z^{(k)}$; under assumptions 1 and 2 this step is efficient. If the two structures have the same POS sequence (i.e., $y^{(k)}(i,t) = z^{(k)}(i,t)$ for all $(i,t)$) then the algorithm returns this solution. Otherwise, simple updates are made to the $u(i,t)$ variables, based on the $y^{(k)}(i,t)$ and $z^{(k)}(i,t)$ values.

In a moment we'll give an example run of the algorithm. First, though, we give an important theorem:

**Theorem 2** *If at any iteration of the algorithm in Figure 2 we have* $y^{(k)}(i,t) = z^{(k)}(i,t)$ *for all* $(i,t)$*, then* $(y^{(k)}, z^{(k)})$ *is a solution to optimization problem 1.*

This theorem is a direct consequence of theorem 5 of this paper.

Thus if we *do* reach agreement between $y^{(k)}$ and $z^{(k)}$, then we are *guaranteed to have an optimal solution to the original problem*. Later in this tutorial we will give empirical results for various NLP problems showing how often, and how quickly, we reach agreement. We will also describe the theory underlying convergence; theory underlying cases where the algorithm doesn't converge; and methods that can be used to "tighten" the algorithm with the goal of achieving convergence.

Next, consider the efficiency of the algorithm. To be concrete, again consider the case where $f(y)$ is defined through a weighted CFG, and $g(z)$ is defined through a finite-state tagger. Each iteration of the algorithm requires decoding under each of these two models. If the number of iterations $k$ is relatively small, the algorithm can be much more efficient than using the construction of Bar-Hillel et al. (1964). As discussed before, assuming a context-free grammar in Chomsky normal form, and a trigram tagger with $T$ tags, the CKY parsing algorithm takes $O(G^3n^3)$ time, and the Viterbi algorithm for tagging takes $O(T^3n)$ time. Thus the total running time for the dual decomposition algorithm is $O(k(G^3n^3 + T^3n))$ where $k$ is the number of iterations required for convergence. In contrast, the construction of Bar-Hillel et al. results in an algorithm with running time of $O(G^3T^6n^3)$. The dual decomposition algorithm results in an *additive* cost for incorporating a tagger (a $T^3n$ term is added into the run time), whereas the construction of Bar-Hillel et al. results in a much more expensive *multiplicative* cost (a $T^6$ term is multiplied into the run time). (Smith & Eisner, 2008, makes a similar observation about additive versus multiplicative costs in the context of belief propagation algorithms for dependency parsing.)

### 4.3 Relationship of the Approach to Section 3

It is easily verified that the approach we have described is an instance of the dual decomposition framework described in section 3.2. The set $\mathcal{Y}$ is the set of all parses for the input sentence; the set $\mathcal{Z}$ is the set of all POS sequences for the input sentence. Each parse tree $y \in \mathbb{R}^d$ is represented as a vector such that $f(y) = y \cdot \theta^{(1)}$ for some $\theta^{(1)} \in \mathbb{R}^d$: there are a number of ways of representing parse trees as vectors, see the work of Rush et al. (2010) for one example. Similarly, each tag sequence $z \in \mathbb{R}^{d'}$ is represented as a vector such that $g(z) = z \cdot \theta^{(2)}$ for some $\theta^{(2)} \in \mathbb{R}^{d'}$. The constraints

$$y(i,t) = z(i,t)$$

for all $(i,t)$ can be encoded through linear constraints

$$Ay + Cz = b$$

for suitable choices of $A$, $C$, and $b$, assuming that the vectors $y$ and $z$ include components $y(i,t)$ and $z(i,t)$ respectively.

### 4.4 An Example Run of the Algorithm

We now give an example run of the algorithm. For simplicity, we will assume that the step size $\delta_k$ is equal to 1 for all iterations $k$. We take the input sentence to be *United flies some large jet*. Initially, the algorithm sets $u(i,t) = 0$ for all $(i,t)$. For our example, decoding with these initial weights leads to the two hypotheses

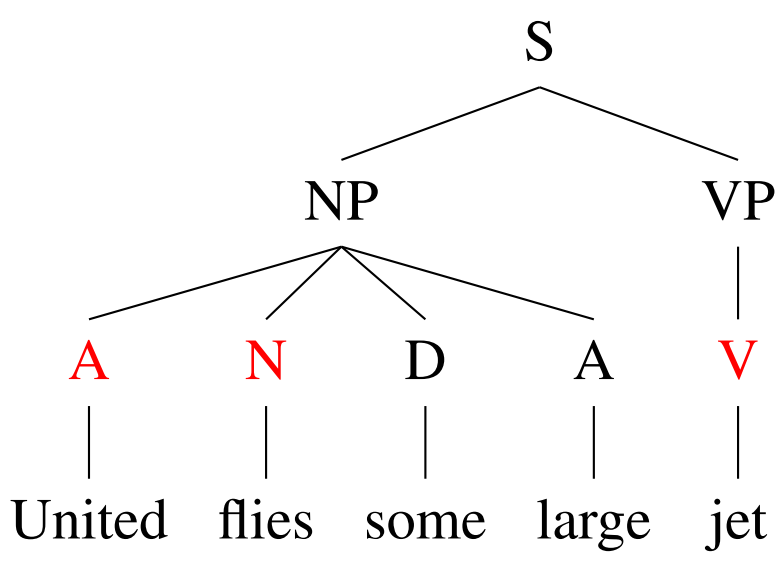

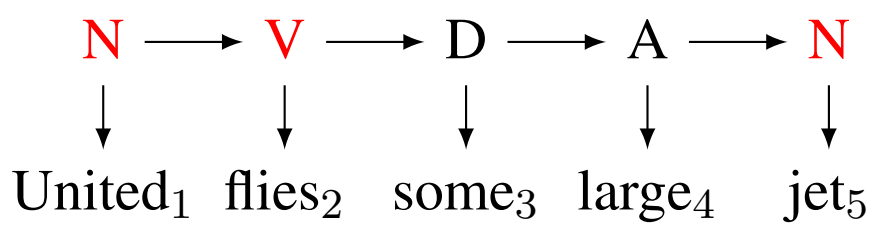

These two structures have different POS tags at three positions, highlighted in red; thus the two structures do not agree. We then update the $u(i, t)$ variables based on these differences, giving new values as follows:

$$u(1, A) = u(2, N) = u(5, V) = -1$$

$$u(1, N) = u(2, V) = u(5, N) = 1$$

Any $u(i, t)$ values not shown still have value $0$. We now decode with these new $u(i, t)$ values, giving structures

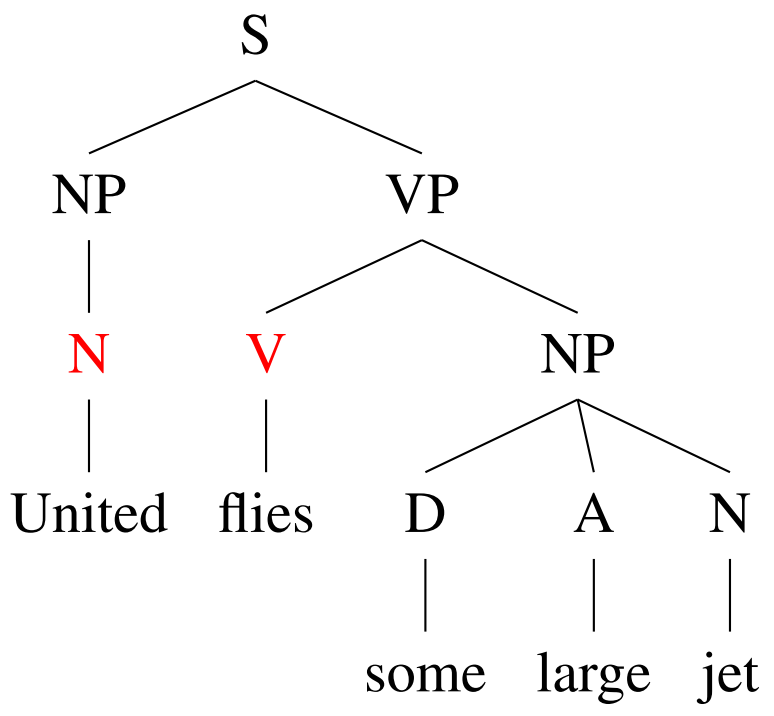

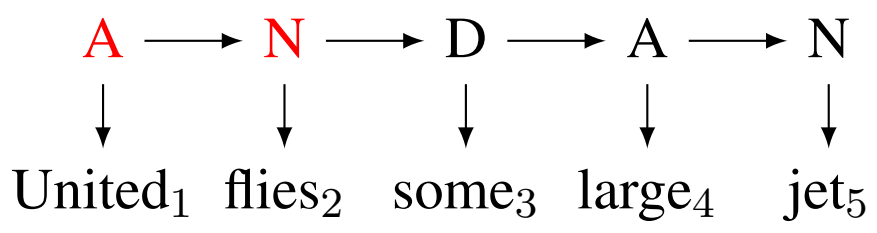

Again, differences between the structures are shown in red. We update the $u(i, t)$ values to obtain new values as follows:

$$u(5, N) = -1$$

$$u(5, V) = 1$$

with all other $u(i, t)$ values being $0$. (Note that the updates reset $u(1, A)$, $u(1, N)$, $u(2, N)$ and $u(2, V)$ back to zero.)

We decode again, with the new $u(i, t)$ values; this time, the two structures are

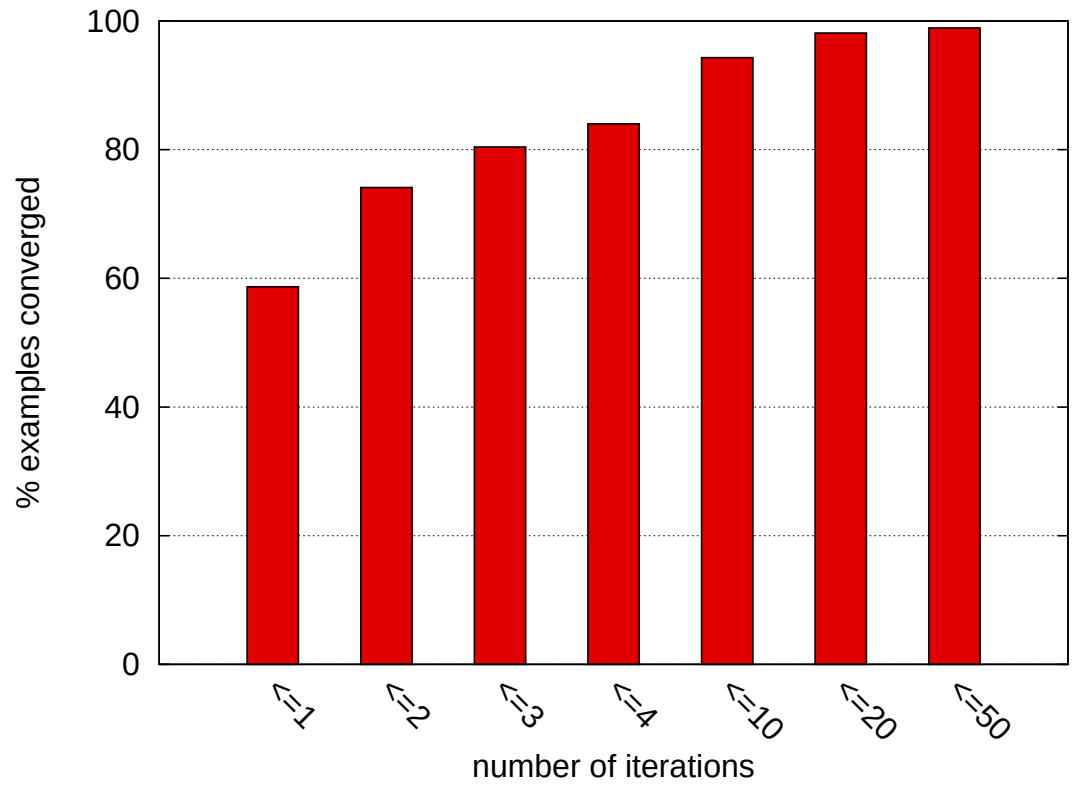

Figure 3: Convergence results from the work of Rush et al. (2010) for integration of a probabilistic parser and a POS tagger, using dual decomposition. We show the percentage of examples where an exact solution is returned by the algorithm, versus the number of iterations of the algorithm.

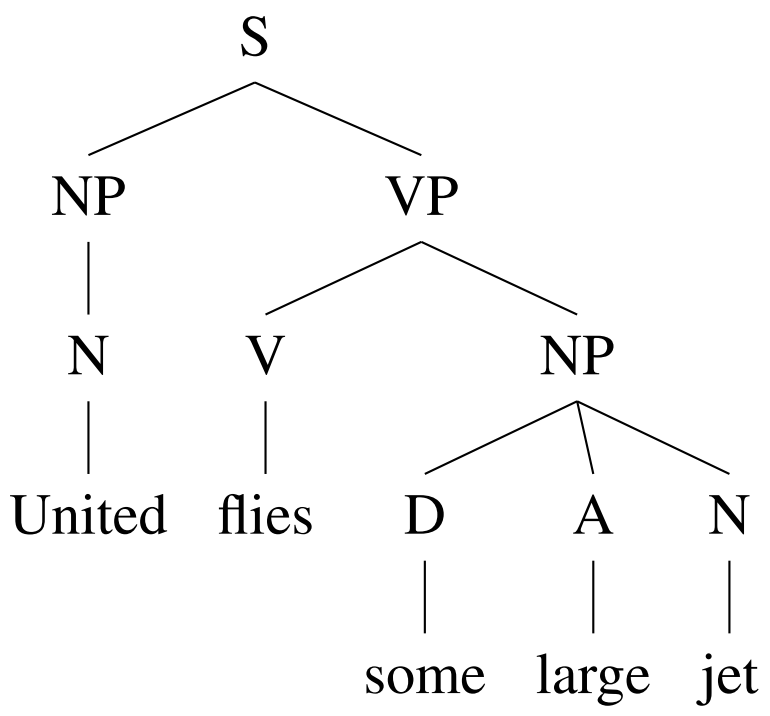

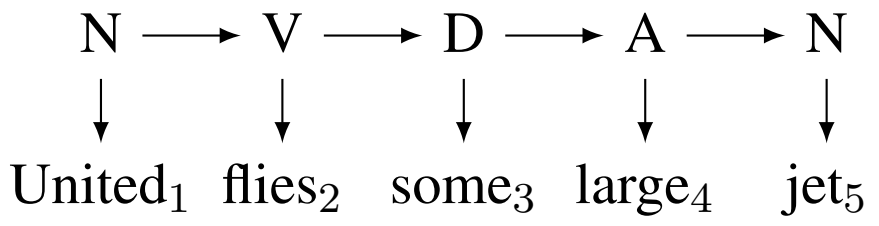

These two structures have identical sequences of POS tags, and the algorithm terminates, with the guarantee that the solutions are optimal.

Rush et al. (2010) describe experiments using this algorithm to integrate the probabilistic parser of Collins (1997) with the POS tagger of Toutanova, Klein, Manning, and Singer (2003). (In these experiments the stepsize $\delta_k$ is not held constant, but is instead set using the strategy described in section 7.2 of this paper.) Figure 3 shows the percentage of cases where exact solutions are returned (we have agreement between $y^{(k)}$ and $z^{(k)}$) versus the number of iterations of the algorithm. The algorithm produces exact solutions on over 99% of all examples. On over 94% of the examples the algorithm returns an exact solution in 10 iterations or fewer. So with these models at least, while the dual decomposition algorithm is not guaranteed to give an exact solution, in this case it is very successful at achieving this goal.

## 5. Formal Properties

We now give some formal properties of the algorithm described in the previous section. We first describe three important theorems regarding the algorithm, and then describe connections between the algorithm and subgradient optimization methods.

### 5.1 Three Theorems

Recall that the problem we are attempting to solve (optimization problem 1) is

$$\underset{y \in \mathcal{Y}, z \in \mathcal{Z}}{\operatorname{argmax}} f(y) + g(z)$$

such that for all $i = 1 \ldots n, t \in \mathcal{T}$,

$$y(i,t) = z(i,t)$$

The first step will be to introduce the Lagrangian for this problem. We introduce a Lagrange multiplier $u(i,t)$ for each equality constraint $y(i,t) = z(i,t)$: we write $u = \{u(i,t) : i \in \{1 \ldots n\}, t \in \mathcal{T}\}$ to denote the vector of Lagrange muliplers. Each Lagrange multiplier can take any positive or negative value. The Lagrangian is

$$L(u,y,z) \quad = \quad f(y) + g(z) + \sum_{i,t} u(i,t) \left( y(i,t) - z(i,t) \right) \tag{12}$$

Note that by grouping the terms that depend on $y$ and $z$, we can rewrite the Lagrangian as

$$L(u,y,z) \quad = \quad \left( f(y) + \sum_{i,t} u(i,t) y(i,t) \right) + \left( g(z) - \sum_{i,t} u(i,t) z(i,t) \right)$$

Having defined the Lagrangian, the *dual objective* is

$$\begin{aligned} L(u) &= \max_{y \in \mathcal{Y}, z \in \mathcal{Z}} L(u,y,z) \\ &= \max_{y \in \mathcal{Y}} \left( f(y) + \sum_{i,t} u(i,t) y(i,t) \right) + \max_{z \in \mathcal{Z}} \left( g(z) - \sum_{i,t} u(i,t) z(i,t) \right) \end{aligned}$$

Under assumptions 1 and 2 described above, the dual value $L(u)$ for any value of $u$ can be calculated efficiently: we simply compute the two $\max$'s, and sum them. Thus the dual decomposes in a very convenient way into two efficiently solvable sub-problems.

Finally, the dual problem is to minimize the dual objective, that is, to find

$$\min_u L(u)$$

We will see shortly that the algorithm in Figure 2 is a subgradient algorithm for minimizing the dual objective.

Define $(y^*, z^*)$ to be the optimal solution to optimization problem 1. The first theorem is as follows:

**Theorem 3** *For any value of* $u$*,*

$$L(u) \geq f(y^*) + g(z^*)$$

Hence $L(u)$ provides an upper bound on the score of the optimal solution. The proof is simple:

*Proof:*

$$L(u) = \max_{y \in \mathcal{Y}, z \in \mathcal{Z}} L(u, y, z) \tag{13}$$

$$\geq \max_{y \in \mathcal{Y}, z \in \mathcal{Z} : y = z} L(u, y, z) \tag{14}$$

$$= \max_{y \in \mathcal{Y}, z \in \mathcal{Z} : y = z} f(y) + g(z) \tag{15}$$

$$= f(y^*) + g(z^*) \tag{16}$$

Here we use the shorthand $y = z$ to state that $y(i,t) = z(i,t)$ for all $(i,t)$. Eq. 14 follows because by adding the constraints $y = z$, we are optimizing over a smaller set of $(y, z)$ pairs, and hence the $\max$ cannot increase. Eq. 15 follows because if $y = z$, we have

$$\sum_{i,t} u(i,t) \left(y(i,t) - z(i,t)\right) = 0$$

and hence $L(u, y, z) = f(y) + g(z)$. Finally, Eq. 16 follows through the definition of $y^*$ and $z^*$. □

The property that $L(u) \geq f(y^*) + g(z^*)$ for any value of $u$ is often referred to as *weak duality*. The value of $\inf_u L(u) - f(y^*) - g(z^*)$ is often referred to as the *duality gap* or the *optimal duality gap* (see for example Boyd & Vandenberghe, 2004).

Note that obtaining an upper bound on $f(y^*) + g(z^*)$ (providing that it is relatively tight) can be a useful goal in itself. First, upper bounds of this form can be used as admissible heuristics for search methods such as A* or branch-and-bound algorithms. Second, if we have some method that generates a potential solution $(y, z)$, we immediately obtain an upper bound on how far this solution is from being optimal, because

$$(f(y^*) + g(z^*)) - (f(y) + g(z)) \leq L(u) - (f(y) + g(z))$$

Hence if $L(u) - (f(y) + g(z))$ is small, then $(f(y^*) + g(z^*)) - (f(y) + g(z))$ must be small. See section 7 for more discussion.

Our second theorem states that the algorithm in Figure 2 successfully converges to $\min_u L(u)$. Hence the algorithm successfully converges to the tightest possible upper bound given by the dual. The theorem is as follows:

**Theorem 4** *Consider the algorithm in Figure 2. For any sequence $\delta_1, \delta_2, \delta_3, \ldots$ such that $\delta_k > 0$ for all $k \geq 1$, and*

$$\lim_{k \to \infty} \delta_k = 0 \quad \textit{and} \quad \sum_{k=1}^{\infty} \delta_k = \infty,$$

*we have*

$$\lim_{k \to \infty} L(u^k) = \min_u L(u)$$

*Proof:* See the work of Shor (1985). See also Appendix A.3. □

Our algorithm is actually a *subgradient method* for minimizing $L(u)$: we return to this point in section 5.2. For now though, the important point is that our algorithm successfully minimizes $L(u)$.

Our final theorem states that if we ever reach agreement during the algorithm in Figure 2, we are guaranteed to have the optimal solution. We first need the following definitions:

**Definition 1** *For any value of $u$, define*

$$y^{(u)} = \underset{y \in \mathcal{Y}}{\operatorname{argmax}} \left( f(y) + \sum_{i,t} u(i,t) y(i,t) \right)$$

*and*

$$z^{(u)} = \underset{z \in \mathcal{Z}}{\operatorname{argmax}} \left( g(z) - \sum_{i,t} u(i,t) z(i,t) \right)$$

The theorem is then:

**Theorem 5** *If $\exists u$ such that*

$$y^{(u)}(i,t) = z^{(u)}(i,t)$$

*for all $i, t$, then*

$$f(y^{(u)}) + g(z^{(u)}) = f(y^*) + g(z^*)$$

*i.e., $(y^{(u)}, z^{(u)})$ is optimal.*

*Proof:* We have, by the definitions of $y^{(u)}$ and $z^{(u)}$,

$$\begin{aligned} L(u) &= f(y^{(u)}) + g(z^{(u)}) + \sum_{i,t} u(i,t)(y^{(u)}(i,t) - z^{(u)}(i,t)) \\ &= f(y^{(u)}) + g(z^{(u)}) \end{aligned}$$

where the second equality follows because $y^{(u)}(i,t) = z^{(u)}(i,t)$ for all $(i,t)$. But $L(u) \geq f(y^*) + g(z^*)$ for all values of $u$, hence

$$f(y^{(u)}) + g(z^{(u)}) \geq f(y^*) + g(z^*)$$

Because $y^*$ and $z^*$ are optimal, we also have

$$f(y^{(u)}) + g(z^{(u)}) \leq f(y^*) + g(z^*)$$

hence we must have

$$f(y^{(u)}) + g(z^{(u)}) = f(y^*) + g(z^*)$$

□

Theorems 4 and 5 refer to quite different notions of convergence of the dual decomposition algorithm. For the remainder of this tutorial, to avoid confusion, we will explicitly use the following terms:

- *d-convergence* (short for "dual convergence") will be used to refer to convergence of the dual decomposition algorithm to the minimum dual value: that is, the property that $\lim_{k \to \infty} L(u^{(k)}) = \min_u L(u)$. By theorem 4, assuming appropriate step sizes in the algorithm, we **always** have d-convergence.

- *e-convergence* (short for "exact convergence") refers to convergence of the dual decomposition algorithm to a point where $y(i,t) = z(i,t)$ for all $(i,t)$. By theorem 5, if the dual decomposition algorithm e-converges, then it is guaranteed to have provided the optimal solution. However, the algorithm is **not** guaranteed to e-converge.

### 5.2 Subgradients

The proof of d-convergence, as defined in theorem 4, relies on the fact that the algorithm in Figure 2 is a subgradient algorithm for minimizing the dual objective $L(u)$. Subgradient algorithms are a generalization of gradient-descent methods; they can be used to minimize convex functions that are non-differentiable. This section describes how the algorithm in Figure 2 is derived as a subgradient algorithm.

Recall that $L(u)$ is defined as follows:

$$\begin{aligned} L(u) &= \max_{y \in \mathcal{Y}, z \in \mathcal{Z}} L(u, y, z) \\ &= \max_{y \in \mathcal{Y}} \left( f(y) + \sum_{i,t} u(i,t) y(i,t) \right) + \max_{z \in \mathcal{Z}} \left( g(z) - \sum_{i,t} u(i,t) z(i,t) \right) \end{aligned}$$

and that our goal is to find $\min_u L(u)$.

First, we note that $L(u)$ has the following properties:

- $L(u)$ is a convex function. That is, for any $u^{(1)} \in \mathbb{R}^d$, $u^{(2)} \in \mathbb{R}^d$, $\lambda \in [0, 1]$,

  $$L(\lambda u^{(1)} + (1 - \lambda) u^{(2)}) \leq \lambda L(u^{(1)}) + (1 - \lambda) L(u^{(2)})$$

  (The proof is simple: see Appendix A.1.)

- $L(u)$ is not differentiable. In fact, it is easily shown that it is a piecewise linear function.

The fact that $L(u)$ is not differentiable means that we cannot use a gradient descent method to minimize it. However, because it is nevertheless a convex function, we can instead use a subgradient algorithm. The definition of a subgradient is as follows:

**Definition 2 (Subgradient)** *A subgradient of a convex function $L : \mathbb{R}^d \to \mathbb{R}$ at $u$ is a vector $\gamma^{(u)}$ such that for all $v \in \mathbb{R}^d$,*

$$L(v) \geq L(u) + \gamma^{(u)} \cdot (v - u)$$

The subgradient $\gamma^{(u)}$ is a tangent at the point $u$ that gives a lower bound to $L(u)$: in this sense it is similar[8] to the gradient for a convex but differentiable function.[9] The key idea in subgradient methods is to use subgradients in the same way that we would use gradients in gradient descent methods. That is, we use updates of the form

$$u' = u - \delta \gamma^{(u)}$$

where $u$ is the current point in the search, $\gamma^{(u)}$ is a subgradient at this point, $\delta > 0$ is a step size, and $u'$ is the new point in the search. Under suitable conditions on the stepsizes $\delta$ (e.g., see theorem 4), these updates will successfully converge to the minimum of $L(u)$.

8. More precisely, for a function $L(u)$ that is convex and differentiable, then its gradient at any point $u$ is a subgradient at $u$.
9. It should be noted, however, that for a given point $u$, there may be more than one subgradient: this will occur, for example, for a piecewise linear function at points where the gradient is not defined.

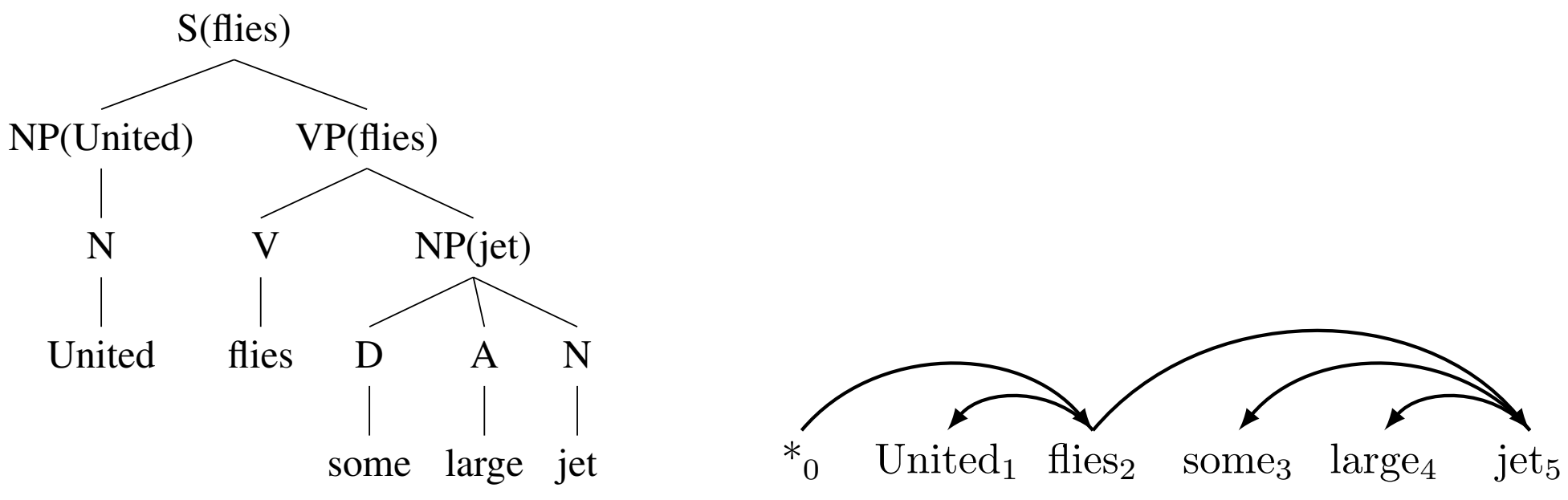

Figure 4: A lexicalized parse tree, and a dependency structure.

So how do we calculate the subgradient for $L(u)$? It turns out that it has a very convenient form. As before (see definition 1), define $y^{(u)}$ and $z^{(u)}$ to be the $\arg\max$'s for the two maximization problems in $L(u)$. If we define the vector $\gamma^{(u)}$ as

$$\gamma^{(u)}(i,t) = y^{(u)}(i,t) - z^{(u)}(i,t)$$

for all $(i,t)$, then it can be shown that $\gamma^{(u)}$ is a subgradient of $L(u)$ at $u$. The updates in the algorithm in Figure 2 take the form

$$u'(i,t) = u(i,t) - \delta(y^{(u)}(i,t) - z^{(u)}(i,t))$$

and hence correspond directly to subgradient updates.

See Appendix A.2 for a proof that the subgradients take this form, and Appendix A.3 for a proof of convergence for the subgradient optimization method.

## 6. Other Examples

In this section we describe other examples of dual decomposition algorithms. Our first example, also from the work of Rush et al. (2010), is a dual decomposition algorithm that combines two parsing models. Our second example, from the work of Komodakis et al. (2007, 2011), is a dual decomposition algorithm for inference in Markov random fields. Finally, we describe the algorithm of Held and Karp (1971) for the traveling salesman problem, and the algorithm of Chang and Collins (2011) for decoding of phrase-based translation models.

### 6.1 Combined Constituency and Dependency Parsing

Rush et al. (2010) describe an algorithm for finding the highest scoring lexicalized context-free parse tree for an input sentence, under a combination of two models: a lexicalized probabilistic context-free grammar, and a discriminative dependency parsing model.

Figure 4 shows an example of a lexicalized context-free tree. We take $\mathcal{Y}$ to be the set of all lexicalized trees for the input sentence, and $f(y)$ to be the score of the tree $y$ under a lexicalized parsing model—specifically, $f(y)$ is the log-probability of $y$ under the model of Collins (1997). Under this model, each lexicalized rule in $y$ receives a score that is a log probability, and the log probability of $y$ is a sum of the log probabilities for the rules that it contains.

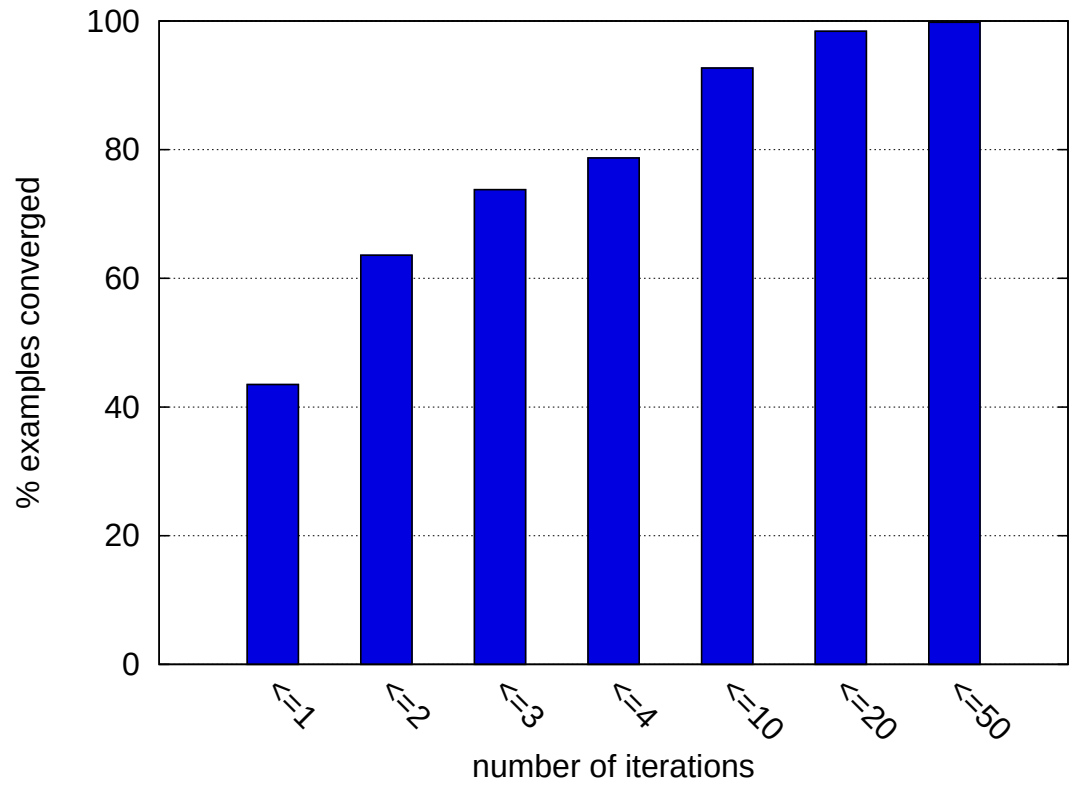

Figure 5: Convergence results from the work of Rush et al. (2010) for integration of a lexicalized probabilistic context-free grammar, and a discriminative dependency parsing model. We show the percentage of examples where an exact solution is returned by the algorithm, versus the number of iterations of the algorithm.

Our second model is a dependency parsing model. An example dependency parse is also shown in Figure 4. The set of all possible dependency parses for the sentence is $\mathcal{Z}$; each parse $z$ receives a score $g(z)$ under the dependency parsing model. We use the discriminative dependency parsing model of Koo, Carreras, and Collins (2008) (see also McDonald, 2006).

For any lexicalized parse tree $y$, there is a mapping to an underlying dependency structure $l(y)$. The decoding problem we consider is to find

$$\operatorname*{argmax}_{y \in \mathcal{Y}} f(y) + g(l(y)) \tag{17}$$

The motivation for this problem is that it will allow us to inject information from the dependency parsing model $g(z)$ into the lexicalized parsing model of Collins (1997); Rush et al. (2010) show that this gives significant improvements in parsing accuracy.

The problem can be again solved exactly using a dynamic programming approach, where a dynamic program is created that is an intersection of the two models (there is a clear analogy to the Bar-Hillel et al. (1964) method for construction of a dynamic program for the intersection of a PCFG and an HMM). However this dynamic program is again relatively inefficient.

We develop a dual decomposition algorithm in a very similar way to before. For any dependency $(i, j)$ where $i \in \{0 \dots n\}$ is the head word (we use 0 to denote the root symbol) and $j \in \{1 \dots n\}$, $j \neq i$, is the modifier, we define $y(i, j) = 1$ if $y$ contains the dependency $(i, j)$, and $y(i, j) = 0$ otherwise. We define similar variables $z(i, j)$ for dependency structures. We can then reformulate the problem in Eq. 17 as:

**Optimization Problem 2** *Find*

$$\operatorname*{argmax}_{y \in \mathcal{Y}, z \in \mathcal{Z}} f(y) + g(z) \tag{18}$$

*such that for all* $(i, j)$, $y(i, j) = z(i, j)$.

A Lagrangian is introduced for this problem, which is very similar to that in Eq. 12, and a subgradient algorithm is used to minimize the resulting dual. We introduce Lagrange multipliers $u(i, j)$ for all dependencies $(i, j)$, whose initial values are $u^{(0)}(i, j) = 0$ for all $i, j$. At each iteration of the algorithm we find

$$y^{(k)} = \underset{y \in \mathcal{Y}}{\operatorname{argmax}} \left( f(y) + \sum_{i,j} u^{(k-1)}(i, j) y(i, j) \right)$$

using a dynamic programming algorithm for lexicalized context-free parsing (a trivial modification of the original algorithm for finding $\operatorname{argmax}_y f(y)$). In addition we find

$$z^{(k)} = \underset{z \in \mathcal{Z}}{\operatorname{argmax}} \left( g(z) - \sum_{i,j} u^{(k-1)}(i, j) z(i, j) \right)$$

using a dynamic programming algorithm for dependency parsing (again, this requires a trivial modification to an existing algorithm). If $y^{(k)}(i, j) = z^{(k)}(i, j)$ for all $(i, j)$ then the algorithm has e-converged, and we are guaranteed to have a solution to optimization problem 2. Otherwise, we perform subgradient updates

$$u^{(k)}(i, j) = u^{(k-1)}(i, j) - \delta_k (y^{(k)}(i, j) - z^{(k)}(i, j))$$

for all $(i, j)$, then go to the next iteration.

Rush et al. (2010) describe experiments with this algorithm. The method e-converges on over 99% of examples, with over 90% of examples e-converging in 10 iterations or less. Figure 5 shows a histogram of the number of examples that have e-converged, versus the number of iterations of the algorithm. The method gives significant gains in parsing accuracy over the model of Collins (1997), and significant gains over a baseline method that simply forces the lexicalized CFG parser to have the same dependency structure as the first-best output from the dependency parser.[10]

### 6.2 The MAP Problem for Pairwise Markov Random Fields

Markov random fields (MRFs), and more generally graphical models, are widely used in machine learning and statistics. The MAP problem in MRFs— the problem of finding the most likely setting of the random variables in an MRF—is an inference problem of central importance. In this section we describe the dual decomposition algorithm from the work of Komodakis et al. (2007, 2011) for finding the MAP solution in pairwise, binary, MRFs. Pairwise MRFs are limited to the case where potential functions consider pairs of random variables, as opposed to larger subsets; however, the generalization of the method to non-pairwise MRFs is straightforward.

A commonly used approach for the MAP problem in MRFs is to use loopy max-product belief propagation. The dual decomposition algorithm has advantages in terms of stronger formal guarantees, as described in section 5.

10. Note that Klein and Manning (2002) describe a method for combination of a dependency parser with a constituent based parser, where the score for an entire structure is again the sum of scores under two models. In this approach an A* algorithm is developed, where admissible estimates within the A* method can be computed efficiently using separate inference under the two models. There are interesting connections between the A* approach and the dual decomposition algorithm described in this section.

The MAP problem is as follows. Assume we have a vector $y$ of variables $y_1, y_2, \ldots, y_n$, where each $y_i$ can take two possible values, 0 or 1 (the generalization to more than two possible values for each variable is straightforward). There are $2^n$ possible settings of these $n$ variables. An MRF assumes an underlying undirected graph $(V, E)$, where $V = \{1 \ldots n\}$ is the set of vertices in the graph, and $E$ is a set of edges. The MAP problem is then to find

$$\underset{y \in \{0,1\}^n}{\operatorname{argmax}}\, h(y) \tag{19}$$

where

$$h(y) = \sum_{\{i,j\} \in E} \theta_{i,j}(y_i, y_j)$$

Here each $\theta_{i,j}(y_i, y_j)$ is a local potential associated with the edge $\{i, j\} \in E$, which returns a real value (positive or negative) for each of the four possible settings of $(y_i, y_j)$.

If the underlying graph $E$ is a tree, the problem in Eq. 19 is easily solved using max-product belief propagation, a form of dynamic programming. In contrast, for general graphs $E$, which may contain loops, the problem is NP-hard. The key insight behind the dual decomposition algorithm will be to decompose the graph $E$ into $m$ trees $T_1, T_2, \ldots, T_m$. Inference over each tree can be performed efficiently; we use Lagrange multipliers to enforce agreement between the inference results for each tree. A subgradient algorithm is used, where at each iteration we first perform inference over each of the trees $T_1, T_2, \ldots, T_m$, and then update the Lagrange multipliers in cases where there are disagreements.

For simplicity, we describe the case where $m = 2$. Assume that the two trees are such that $T_1 \subset E$, $T_2 \subset E$, and $T_1 \cup T_2 = E$.[11] Thus each of the trees contains a subset of the edges in $E$, but together the trees contain all edges in $E$. Assume that we define potential functions $\theta^{(1)}_{i,j}$ for $(i, j) \in T_1$ and $\theta^{(2)}_{i,j}$ for $(i, j) \in T_2$ such that

$$\sum_{\{i,j\} \in E} \theta_{i,j}(y_i, y_j) = \sum_{\{i,j\} \in T_1} \theta^{(1)}_{i,j}(y_i, y_j) + \sum_{\{i,j\} \in T_2} \theta^{(2)}_{i,j}(y_i, y_j)$$

This is easy to do: for example, define

$$\theta^{m}_{i,j}(y_i, y_j) = \frac{\theta_{i,j}(y_i, y_j)}{\#(i, j)}$$

for $m = 1, 2$ where $\#(i, j)$ is 2 if the edge $\{i, j\}$ appears in both trees, 1 otherwise.

We can then define a new problem that is equivalent to the problem in Eq. 19:

**Optimization Problem 3** *Find*

$$\underset{y \in \{0,1\}^n, z \in \{0,1\}^n}{\operatorname{argmax}} \sum_{\{i,j\} \in T_1} \theta^{(1)}_{i,j}(y_i, y_j) + \sum_{\{i,j\} \in T_2} \theta^{(2)}_{i,j}(z_i, z_j)$$

*such that* $y_i = z_i$ *for* $i = 1 \ldots n$.

11. It may not always be possible to decompose a graph $E$ into just 2 trees in this way. Komodakis et al. (2007, 2011) describe an algorithm for the general case of more than 2 trees.

Note the similarity to our previous optimization problems. Our goal is to find a pair of structures, $y \in \{0,1\}^n$ and $z \in \{0,1\}^n$. The objective function can be written as

$$f(y) + g(z)$$

where

$$f(y) = \sum_{\{i,j\} \in T_1} \theta^{(1)}_{i,j}(y_i, y_j)$$

and

$$g(z) = \sum_{\{i,j\} \in T_2} \theta^{(2)}_{i,j}(z_i, z_j)$$

We have a set of constraints, $y_i = z_i$ for $i = 1 \ldots n$, which enforce agreement between $y$ and $z$.

We then proceed as before—we define a Lagrangian with a Lagrange multiplier $u_i$ for each constraint:

$$L(u, y, z) = \sum_{\{i,j\} \in T_1} \theta^{(1)}_{i,j}(y_i, y_j) + \sum_{\{i,j\} \in T_2} \theta^{(2)}_{i,j}(z_i, z_j) + \sum_{i=1}^{n} u_i(y_i - z_i)$$

We then minimize the dual

$$L(u) = \max_{y,z} L(u, y, z)$$

using a subgradient algorithm. The algorithm is initialized with $u^{(0)}_i = 0$ for $i = 1 \ldots n$. At each iteration of the algorithm we find

$$y^{(k)} = \operatorname*{argmax}_{y \in \{0,1\}^n} \left( \sum_{\{i,j\} \in T_1} \theta^{(1)}_{i,j}(y_i, y_j) + \sum_i u^{(k-1)}_i y_i \right)$$

and

$$z^{(k)} = \operatorname*{argmax}_{z \in \{0,1\}^n} \left( \sum_{\{i,j\} \in T_2} \theta^{(2)}_{i,j}(z_i, z_j) - \sum_i u^{(k-1)}_i z_i \right)$$

These steps can be achieved efficiently, because $T_1$ and $T_2$ are trees, hence max-product belief propagation produces an exact answer. (The Lagrangian terms $\sum_i u^{(k-1)}_i y_i$ and $\sum_i u^{(k-1)}_i z_i$ are easily incorporated.) If $y^{(k)}_i = z^{(k)}_i$ for all $i$ then the algorithm has e-converged, and we are guaranteed to have a solution to optimization problem 3. Otherwise, we perform subgradient updates of the form

$$u^{(k)}_i = u^{(k-1)}_i - \delta_k (y^{(k)}_i - z^{(k)}_i)$$

for $i = \{1 \ldots n\}$, then go to the next iteration. Intuitively, these updates will bias the two inference problems towards agreement with each other.

Komodakis et al. (2007, 2011) show good experimental results for the method. The algorithm has some parallels to max-product belief propagation, where the $u_i$ values can be interpreted as "messages" being passed between sub-problems.

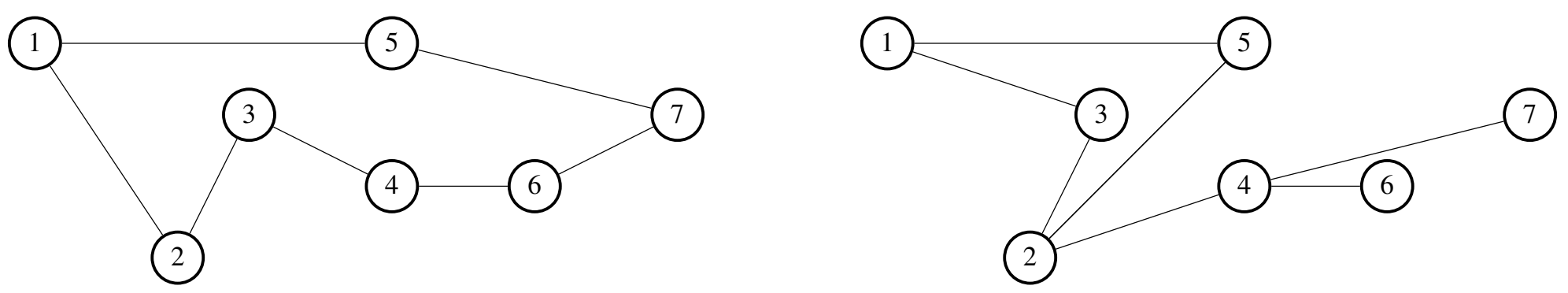

Figure 6: An illustration of the approach of Held and Karp (1971). On the left is a tour of the vertices $1 \dots 7$. On the right is a 1-tree over the vertices $1 \dots 7$. A 1-tree consists of a tree over vertices $2 \dots 7$, together with 2 additional edges that include vertex 1. The tour on the left is also a 1-tree (in fact every tour is also a 1-tree).

### 6.3 The Held and Karp Algorithm for TSPs

Our next example is the approach of Held and Karp (1971) for traveling salesman problems (TSPs), which is notable for being the original paper on Lagrangian relaxation. This algorithm is not an instance of dual decomposition. Instead of leveraging two or more combinatorial algorithms, in combination with agreement constraints, it makes use of a *single* combinatorial algorithm, together with a set of linear constraints that are again incorporated using Lagrange multipliers. While the use of two or more combinatorial algorithms, as seen in dual decomposition, is a very useful technique, broadening our scope to algorithms that make use of a single combinatorial algorithm will be very useful. For NLP decoding algorithms that leverage a single combinatorial algorithm, see the algorithm of Chang and Collins (2011) for decoding of phrase-based translation models (we describe this algorithm in the next section), and the algorithm of Rush and Collins (2011) for decoding of syntax-based translation models.

A TSP is defined as follows. We have an undirected graph $(V, E)$ with vertices $V = \{1, 2, \dots, n\}$, and edges $E$. Each edge $e \in E$ has a score $\theta_e \in \mathbb{R}$. Any subset of the edges $E$ can be represented as a vector $y = \{y_e : e \in E\}$, where $y_e = 1$ if the edge is in the subset, and $y_e = 0$ otherwise. Thus $y$ is a vector in $\{0,1\}^{|E|}$. A *tour* of the graph is a subset of the edges that corresponds to a path through the graph that begins and ends at the same vertex, and includes every other vertex exactly once. See Figure 6 for an example of a tour. We use $\mathcal{Y} \subset \{0,1\}^{|E|}$ to denote the set of all possible tours. The traveling salesman problem is to find

$$\operatorname*{argmax}_{y \in \mathcal{Y}} \sum_{e \in E} y_e \theta_e$$

This problem is well-known to be NP-hard.[12]

A key idea in the work of Held and Karp (1971) is that of a *1-tree*, which, like a tour, is a subset of $E$. Held and Karp define a 1-tree as follows:

> *A 1-tree consists of a tree on the vertex set* $\{2, 3, \dots, n\}$*, together with two distinct edges at vertex 1... Thus, a 1-tree has a single cycle, this cycle contains vertex 1, and vertex 1 always has degree two.*

12. In many presentations of the traveling salesman problem the goal is to find a *minimum* cost tour: for consistency with the rest of this tutorial our presentation considers the maximization problem, which is equivalent.

Figure 6 shows an example 1-tree. We define $\mathcal{Y}'$ to be the set of all possible 1-trees. It follows that $\mathcal{Y}$ is a subset of $\mathcal{Y}'$, because every tour is also a 1-tree.

Crucially, it is possible to find

$$\operatorname*{argmax}_{y \in \mathcal{Y}'} \sum_{e \in E} y_e \theta_e$$

using an efficient algorithm. In the first step, we find the maximum scoring spanning tree over the vertices $\{2, 3, \ldots, n\}$, using a maximum spanning tree algorithm. In the second step, we add the two highest scoring edges that include vertex 1. It is simple to show that the resulting 1-tree is optimal. Thus while search over the set $\mathcal{Y}$ is NP-hard, search over the larger set $\mathcal{Y}'$ can be performed easily. The Lagrangian relaxation algorithm will explicitly leverage this observation.

Next, we note that

$$\mathcal{Y} = \{y : y \in \mathcal{Y}', \text{and for all } i \in \{1, 2, \ldots, n\}, \textstyle\sum_{e:i \in e} y_e = 2\}$$

Each constraint of the form

$$\sum_{e:i \in e} y_e = 2 \tag{20}$$

corresponds to the property that the $i$'th vertex should have exactly two incident edges. Thus if we add the constraint that each vertex has exactly two incident edges, we go from the set of 1-trees to the set of tours. Constraints of the form in Eq. 20 are linear in the $y_e$ variables, and are therefore easily incorporated into a Lagrangian.

Held and Karp introduce the following optimization problem:

**Optimization Problem 4** *Find*

$$\operatorname*{argmax}_{y \in \mathcal{Y}'} \sum_{e \in E} y_e \theta_e$$

*such that for all* $i \in \{1, 2, \ldots, n\}$, $\sum_{e:i \in e} y_e = 2$.

It is clear that this is equivalent to finding the highest scoring tour in the graph.

As before, we deal with the equality constraints using Lagrange multipliers. Define the Lagrange multipliers to be a vector $u = \{u_i : i \in \{1 \ldots n\}\}$. The Lagrangian is

$$L(u, y) = \sum_{e \in E} y_e \theta_e + \sum_{i=1}^{n} u_i \left( \sum_{e:i \in e} y_e - 2 \right)$$

and the dual objective is

$$L(u) = \max_{y \in \mathcal{Y}'} L(u, y)$$

The subgradient algorithm takes the following form. Initially we set $u_i^{(0)} = 0$ for $i = 1 \ldots n$. At each iteration we find

$$y^{(k)} = \operatorname*{argmax}_{y \in \mathcal{Y}'} \left( \sum_{e \in E} y_e \theta_e + \sum_{i=1}^{n} u_i^{(k-1)} \left( \sum_{e:i \in e} y_e - 2 \right) \right) \tag{21}$$

If the constraints are satisfied, i.e., if for all $i$

$$\sum_{e:i \in e} y_e^{(k)} = 2$$

then the algorithm terminates, with a guarantee that the structure $y^{(k)}$ is the solution to optimization problem 4. Otherwise, a subgradient step is used to modify the Lagrange multipliers. It can be shown that the subgradient of $L(u)$ at $u$ is the vector $g^{(u)}$ defined as

$$g^{(u)}(i) = \sum_{e:i\in e} y_e^{(u)} - 2$$

where $y^{(u)} = \text{argmax}_{y\in\mathcal{Y}'} L(u, y)$. Thus the subgradient step is for all $i \in \{1 \dots n\}$,

$$u_i^{(k)} = u_i^{(k-1)} - \delta_k \left( \sum_{e:i\in e} y_e^{(k)} - 2 \right) \tag{22}$$

Note that the problem in Eq. 21 is easily solved. It is equivalent to finding

$$y^{(k)} = \operatorname*{argmax}_{y\in\mathcal{Y}'} \sum_{e\in E} y_e \theta'_e$$

with modified edge weights $\theta'_e$: for an edge $e = \{i, j\}$, we define

$$\theta'_e = \theta_e + u_i^{(k-1)} + u_j^{(k-1)}$$

Hence the new edge weights incorporate the Lagrange multipliers for the two vertices in the edge.

The subgradient step in Eq. 22 has a clear intuition. For vertices with greater than 2 incident edges in $y^{(k)}$, the value of the Lagrange multiplier $u_i$ is decreased, which will have the effect of penalising any edges including vertex $i$. Conversely, for vertices with fewer than 2 incident edges, $u_i$ will increase, and edges including that vertex will be preferred. The algorithm manipulates the $u_i$ values in an effort to enforce the constraints that each vertex has exactly two incident edges.

We note that there is a qualitative difference between this example and our previous algorithms. Our previous algorithms had employed two sets of structures $\mathcal{Y}$ and $\mathcal{Z}$, two optimization problems, and equality constraints enforcing agreement between the two structures. The TSP relaxation instead involves a single set $\mathcal{Y}$. The two approaches are closely related, however, and similar theorems apply to the TSP method (the proofs are trivial modifications of the previous proofs). We have

$$L(u) \geq \sum_{e\in E} y_e^* \theta_e$$

for all $u$, where $y^*$ is the optimal tour. Under appropriate step sizes for the subgradient algorithm, we have

$$\lim_{k\to\infty} L(u^{(k)}) = \min_u L(u)$$

Finally, if we ever find a structure $y^{(k)}$ that satisfies the linear constraints, then the algorithm has e-converged, and we have a guaranteed solution to the traveling salesman problem.

### 6.4 Phrase-Based Translation

We next consider a Lagrangian relaxation algorithm, described in the work of Chang and Collins (2011), for decoding of phrase-based translation models (Koehn, Och, & Marcu, 2003). The input to a phrase-based translation model is a source-language sentence with $n$ words, $x = x_1 \dots x_n$. The output is a sentence in the target language. The examples in this section will use German as the source language, and English as the target language. We will use the German sentence wir müssen auch diese kritik ernst nehmen as a running example.

A key component of a phrase-based translation model is a phrase-based lexicon, which pairs sequences of words in the source language with sequences of words in the target language. For example, lexical entries that are relevent to the German sentence shown above include

(wir müssen, we must)
(wir müssen auch, we must also)
(ernst, seriously)

and so on. Each phrase entry has an associated score, which can take any value in the reals.

We introduce the following notation. A *phrase* is a tuple $(s, t, e)$, signifying that the subsequence $x_s \dots x_t$ in the source language sentence can be translated as the target-language string $e$, using an entry from the phrase-based lexicon. For example, the phrase $(1, 2,$ we must$)$ would specify that the sub-string $x_1 \dots x_2$ can be translated as we must. Each phrase $p = (s, t, e)$ receives a score $\theta(p) \in \mathbb{R}$ under the model. For a given phrase $p$, we will use $s(p)$, $t(p)$ and $e(p)$ to refer to its three components. We will use $\mathcal{P}$ to refer to the set of all possible phrases for the input sentence $x$.

A *derivation* $y$ is then a finite sequence of phrases, $p_1, p_2, \dots p_L$. The length $L$ can be any positive integer value. For any derivation $y$ we use $e(y)$ to refer to the underlying translation defined by $y$, which is derived by concatenating the strings $e(p_1), e(p_2), \dots e(p_L)$. For example, if

$$y = \text{(1, 3, we must also), (7, 7, take), (4, 5, this criticism), (6, 6, seriously)} \tag{23}$$

then

$$e(y) = \text{we must also take this criticism seriously}$$

The score for any derivation $y$ is then defined as

$$h(y) = g(e(y)) + \sum_{p \in y} \theta(p)$$

where $g(e(y))$ is the score (log-probability) for $e(y)$ under an n-gram language model.

The set $\mathcal{Y}$ of valid derivations is defined as follows. For any derivation $y$, we define $y(i)$ for $i = 1 \dots n$ to be the number of times that source word $i$ is translated in the derivation. More formally,

$$y(i) = \sum_{p \in y} [[s(p) \leq i \leq t(p)]]$$

where $[[\pi]]$ is 1 if the statement $\pi$ is true, 0 otherwise. The set of valid derivations is then

$$\mathcal{Y} = \{y \in \mathcal{P}^* : \text{for } i = 1 \dots n, \, y(i) = 1\}$$

where $\mathcal{P}^*$ is the set of finite length sequences of phrases. Thus for a derivation to be valid, each source-language word must be translated exactly once. Under this definition, the derivation in Eq. 23 is valid. The decoding problem is then to find

$$y^* = \operatorname*{argmax}_{y \in \mathcal{Y}} h(y) \tag{24}$$

This problem is known to be NP-hard. Some useful intuition is as follows. A dynamic programming approach for this problem would need to keep track of a bit-string of length $n$ specifying which of the $n$ source language words have or haven't been translated at each point in the dynamic program. There are $2^n$ such bit-strings, resulting in the dynamic program having an exponential number of states.

We now describe the Lagrangian relaxation algorithm. As before, the key idea will be to define a set $\mathcal{Y}'$ such that $\mathcal{Y}$ is a subset of $\mathcal{Y}'$, and such that

$$\operatorname*{argmax}_{y \in \mathcal{Y}'} h(y) \tag{25}$$

can be found efficiently. We do this by defining

$$\mathcal{Y}' = \{y \in \mathcal{P}^* : \textstyle\sum_{i=1}^n y(i) = n\}$$

Thus derivations in $\mathcal{Y}'$ satisfy the weaker constraint that the total number of source words translated is exactly $n$: we have dropped the $y(i) = 1$ constraints. As one example, the following derivation is a member of $\mathcal{Y}'$, but is not a member of $\mathcal{Y}$:

$$y = \text{(1, 3, we must also), (1, 2, we must), (3, 3, also), (6, 6, seriously)} \tag{26}$$

In this case we have $y(1) = y(2) = y(3) = 2$, $y(4) = y(5) = y(7) = 0$, and $y(6) = 1$. Hence some words are translated more than once, and some words are translated 0 times.

Under this definition of $\mathcal{Y}'$, the problem in Eq. 25 can be solved efficiently, using dynamic programming. In contrast to the dynamic program for Eq. 24, which keeps track of a bit-string of length $n$, the new dynamic program merely needs to keep track of how many source language words have been translated at each point in the search.

We proceed as follows. Note that

$$\mathcal{Y} = \{y : y \in \mathcal{Y}', \text{and for } i = 1 \dots n, \, y(i) = 1\}$$

We introduce a Lagrange multiplier $u(i)$ for each constraint $y(i) = 1$. The Lagrangian is

$$L(u, y) = h(y) + \sum_{i=1}^{n} u(i) \left(y(i) - 1\right)$$

The subgradient algorithm is as follows. Initially we set $u^{(0)}(i) = 0$ for all $i$. At each iteration we find

$$y^{(k)} = \operatorname*{argmax}_{y \in \mathcal{Y}'} L(u^{(k-1)}, y) \tag{27}$$

and perform the subgradient step

$$u^{(k)}(i) = u^{(k-1)}(i) - \delta_k (y^{(k)}(i) - 1) \tag{28}$$

If at any point we have $y^{(k)}(i) = 1$ for $i = 1 \dots n$, then we are guaranteed to have the optimal solution to the original decoding problem.

The problem in Eq. 27 can be solved efficiently, because

$$\begin{aligned}
& \operatorname*{argmax}_{y \in \mathcal{Y}'} L(u^{(k-1)}, y) \\
= \; & \operatorname*{argmax}_{y \in \mathcal{Y}'} \left( g(e(y)) + \sum_{p \in y} \theta(p) + \sum_{i=1}^{n} u^{(k-1)}(i)\,(y(i) - 1) \right) \\
= \; & \operatorname*{argmax}_{y \in \mathcal{Y}'} \left( g(e(y)) + \sum_{p \in y} \theta'(p) \right)
\end{aligned}$$

where

$$\theta'(p) = \theta(p) + \sum_{i=s(p)}^{t(p)} u^{(k-1)}(i)$$

Thus we have new phrase scores, $\theta'(p)$, which take into account the Lagrange multiplier values for positions $s(p) \ldots t(p)$. The subgradient step in Eq. 28 has a clear intuition. For any source language word $i$ that is translated more than once, the associated Lagrange multiplier $u(i)$ will decrease, causing phrases including word $i$ to be penalised at the next iteration. Conversely, any word translated 0 times will have its Lagrange multiplier increase, causing phrases including that word to be preferred at the next iteration. The subgradient method manipulates the $u(i)$ values in an attempt to force each source-language word to be translated exactly once.

The description we have given here is a sketch: Chang and Collins (2011) describe details of the method, including a slightly more involved dynamic program that gives a tighter relaxation than the method we have described here, and a tightening method that incrementally adds constraints when the method does not initially e-converge. The method is successful in recovering exact solutions under a phrase-based translation model, and is far more efficient than alternative approaches based on general-purpose integer linear programming solvers.

## 7. Practical Issues

This section reviews various practical issues that arise with dual decomposition algorithms. We describe diagnostics that can be used to track progress of the algorithm in minimizing the dual, and in providing a primal solution; we describe methods for choosing the step sizes, $\delta_k$, in the algorithm; and we describe heuristics that can be used in cases where the algorithm does not provide an exact solution. We will continue to use the algorithm from section 4 as a running example, although our observations are easily generalized to other Lagrangian relaxation algorithms.

The first thing to note is that each iteration of the algorithm produces a number of useful terms, in particular:

- The solutions $y^{(k)}$ and $z^{(k)}$.
- The current dual value $L(u^{(k)})$ (which is equal to $L(u^{(k)}, y^{(k)}, z^{(k)})$.

In addition, in cases where we have a function $l : \mathcal{Y} \to \mathcal{Z}$ that maps each structure $y \in \mathcal{Y}$ to a structure $l(y) \in \mathcal{Z}$, we also have

- A primal solution $y^{(k)}, l(y^{(k)})$.

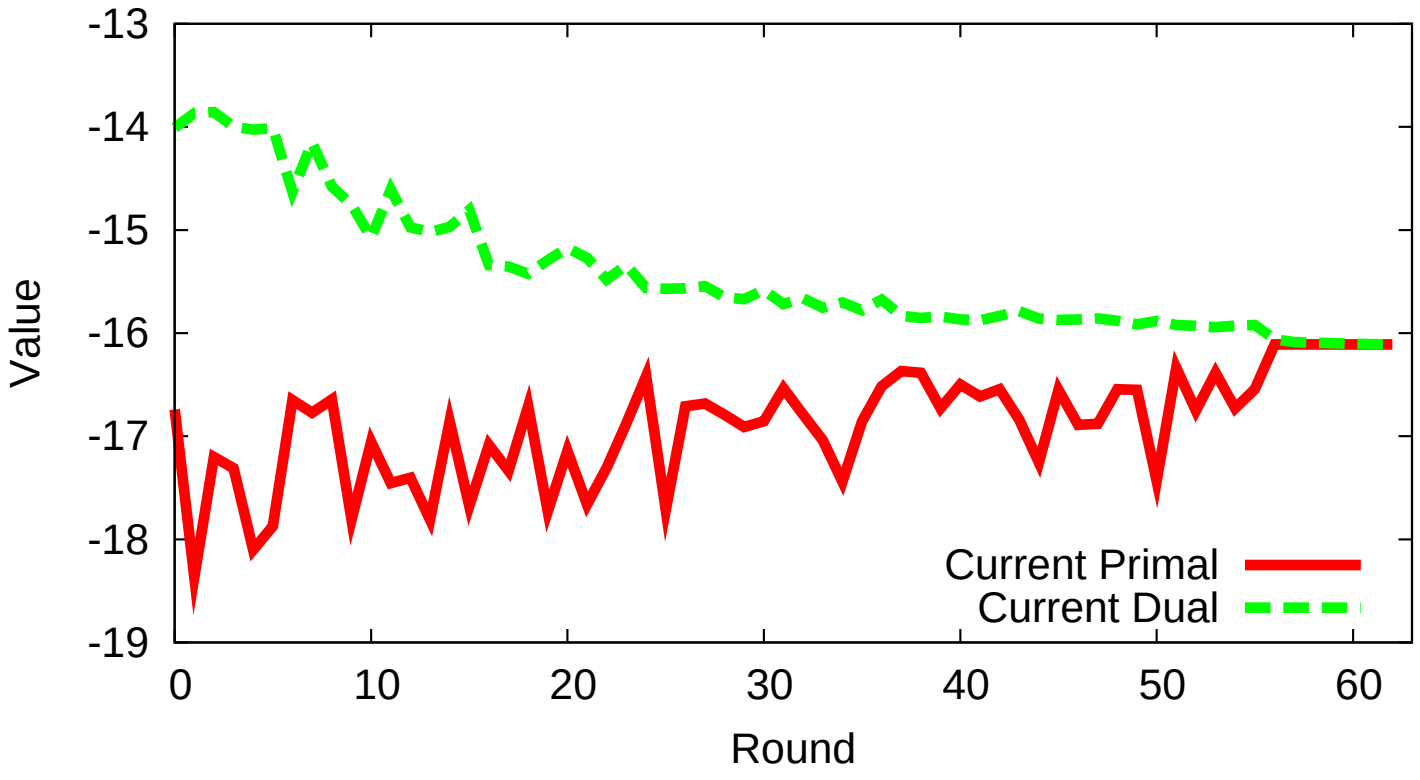

Figure 7: Graph showing the dual value $L(u^{(k)})$ and primal value $f(y^{(k)}) + g(l(y^{(k)}))$, versus iteration number $k$, for the subgradient algorithm on a translation example from the work of Rush and Collins (2011).

- A primal value $f(y^{(k)}) + g(l(y^{(k)}))$.

By a "primal solution" we mean a pair $(y, z)$ that satisfies all constraints in the optimization problem. For example, in optimization problem 1 (the combined HMM and PCFG problem from section 4) a primal solution has the properties that $y \in \mathcal{Y}$, $z \in \mathcal{Z}$, and $y(i,t) = z(i,t)$ for all $(i,t)$.

As one example, in the algorithm in Figure 2, at each iteration we produce a parse tree $y^{(k)}$. It is simple to recover the POS sequence $l(y^{(k)})$ from the parse tree, and to calculate the score $f(y^{(k)})+g(l(y^{(k)}))$ under the combined model. Thus even if $y^{(k)}$ and $z^{(k)}$ disagree, we can still use $y^{(k)}, l(y^{(k)})$ as a potential primal solution. This ability to recover a primal solution from the value $y^{(k)}$ does not always hold—but in cases where it does hold, it is very useful. It will allow us, for example, to recover an approximate solution in cases where the algorithm hasn't e-converged to an exact solution.

We now describe how the various items described above can be used in practical applications of the algorithm.

### 7.1 An Example Run of the Algorithm

Figure 7 shows a run of the subgradient algorithm for the decoding approach for machine translation described in the work of Rush and Collins (2011). The behavior is typical of cases where the algorithm e-converges to an exact solution. We show the dual value $L(u^{(k)})$ at each iteration, and the value for $f(y^{(k)}) + g(l(y^{(k)}))$. A few important points are as follows:

- Because $L(u)$ provides an upper bound on $f(y^*) + g(z^*)$ for any value of $u$, we have

  $$L(u^{(k)}) \geq f(y^{(k)}) + g(l(y^{(k)}))$$

  at every iteration.

- On this example we have e-convergence to an exact solution, at which point we have

  $$L(u^{(k)}) = f(y^{(k)}) + g(z^{(k)}))$$

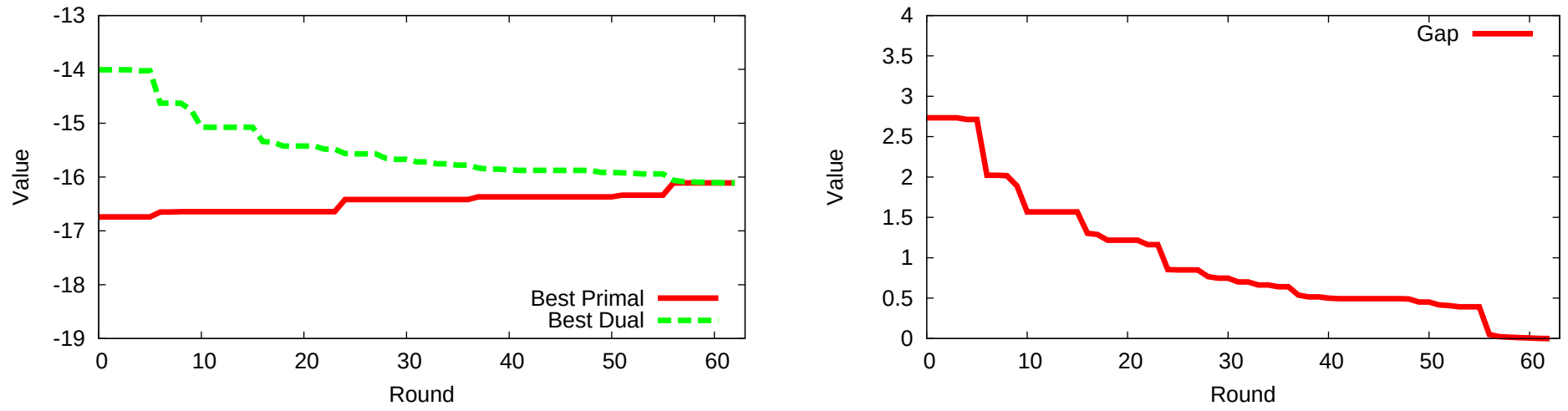

Figure 8: The graph on the left shows the best dual value $L_k^*$ and the best primal value $p_k^*$, versus iteration number $k$, for the subgradient algorithm on a translation example from the work of Rush and Collins (2011). The graph on the right shows $L_k^* - p_k^*$ plotted against $k$.

with $(y^{(k)}, z^{(k)})$ guaranteed to be optimal (and in addition, with $z^{(k)} = l(y^{(k)})$).

- The dual values $L(u^{(k)})$ are not monotonically decreasing—that is, for some iterations we have

$$L(u^{(k+1)}) > L(u^{(k)})$$

even though our goal is to minimize $L(u)$. This is typical: subgradient algorithms are not in general guaranteed to give monotonically decreasing dual values. However, we do see that for most iterations the dual decreases—this is again typical.

- Similarly, the primal value $f(y^{(k)}) + g(z^{(k)})$ fluctuates (goes up and down) during the course of the algorithm.

The following quantities can be useful in tracking progress of the algorithm at the $k$'th iteration:

- $L(u^{(k)}) - L(u^{(k-1)})$ is the change in the dual value from one iteration to the next. We will soon see that this can be useful when choosing the step size for the algorithm (if this value is positive, it may be an indication that the step size should decrease).

- $L_k^* = \min_{k' \leq k} L(u^{(k')})$ is the best dual value found so far. It gives us the tightest upper bound on $f(y^*) + g(z^*)$ that we have after $k$ iterations of the algorithm.

- $p_k^* = \max_{k' \leq k} f(y^{(k')}) + g(l(y^{(k')}))$ is the best primal value found so far.

- $L_k^* - p_k^*$ is the gap between the best dual and best primal solution found so far by the algorithm. Because $L_k^* \geq f(y^*) + g(z^*) \geq p_k^*$, we have

$$L_k^* - p_k^* \geq f(y^*) + g(z^*) - p_k^*$$

hence the value for $L_k^* - p_k^*$ gives us an upper bound on the difference between $f(y^*) + g(z^*)$ and $p_k^*$. If $L_k^* - p_k^*$ is small, we have a guarantee that we have a primal solution that is close to being optimal.

Figure 8 shows a plot of $L_k^*$ and $p_k^*$ versus the number of iterations $k$ for our previous example, and in addition shows a plot of the gap $L_k^* - p_k^*$. These graphs are, not surprisingly, much smoother than the graph in Figure 7. In particular we are guaranteed that the values for $L_k^*$ and $p_k^*$ are monotonically decreasing and increasing respectively.

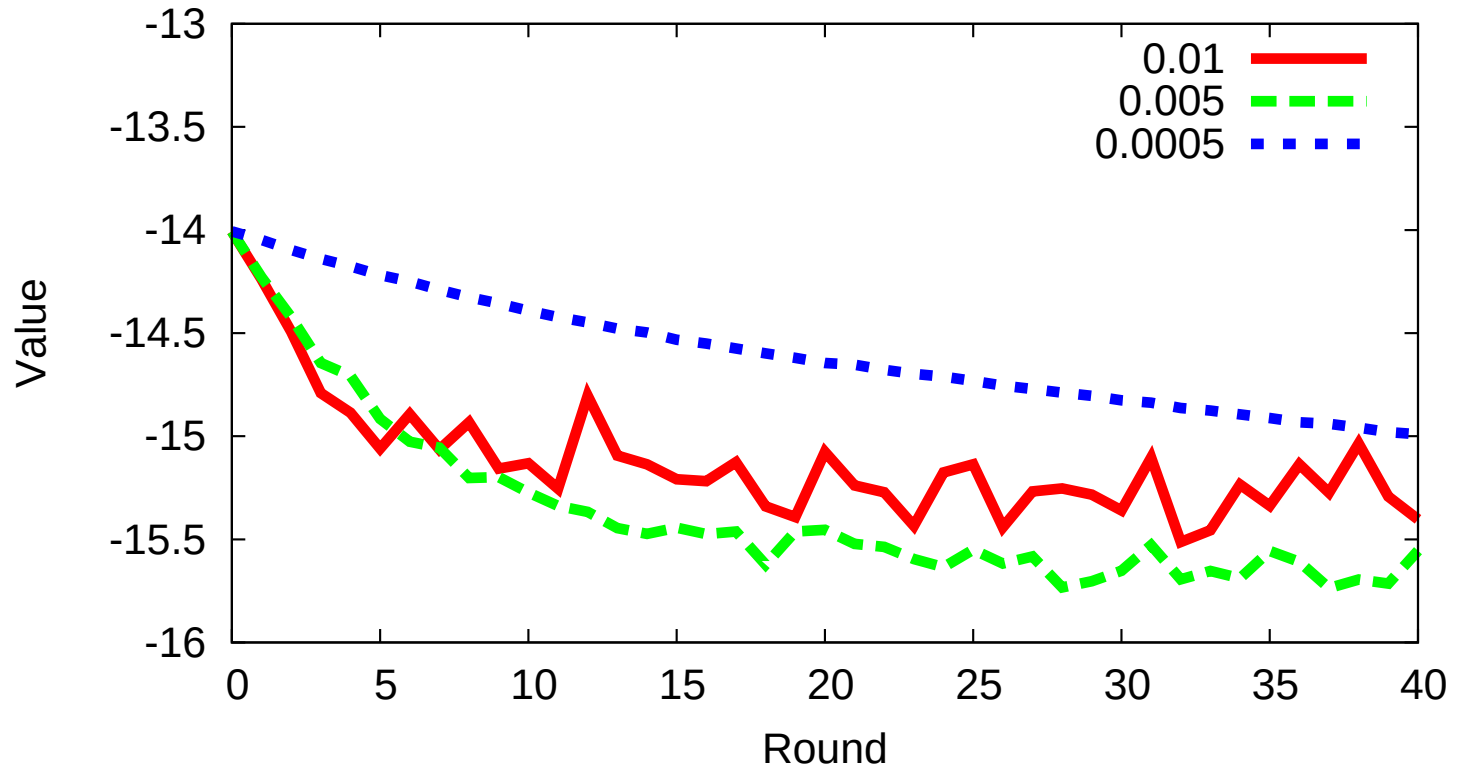

Figure 9: Graph showing the dual value $L(u^{(k)})$ versus the number of iterations $k$, for different fixed step sizes.

### 7.2 Choice of the Step Sizes $\delta_k$

Figure 9 shows convergence of the algorithm for various choices of step size, where we have chosen to keep the stepsize constant at each iteration. We immediately see a potential dilemma arising. With too small a step size ($\delta = 0.0005$), convergence is smooth—the dual value is monotonically decreasing—but convergence is slow. With too large a step size ($\delta = 0.01$), convergence is much faster in the initial phases of the algorithm, but the dual then fluctuates quite erratically. In practice it is often very difficult to choose a constant step size that gives good convergence properties in both early and late iterations of the algorithm.

Instead, we have found that we often find improved convergence properties with a choice of step size $\delta_k$ that decreases with increasing $k$. One possibility is to use a definition such as $\delta_k = c/k$ or $\delta_k = c/\sqrt{k}$ where $c > 0$ is a constant. However these definitions can decrease the step size too rapidly—in particular, they decrease the step size at all iterations, even in cases where progress is being made in decreasing the dual value. In many cases we have found that a more effective definition is

$$\delta_k = \frac{c}{t+1}$$

where $c > 0$ is again a constant, and $t < k$ is the number of iterations prior to $k$ where the dual value increases rather than decreases (i.e., the number of cases for $k' \leq k$ where $L(u^{(k')}) > L(u^{(k'-1)})$). Under this definition the step size decreases only when the dual value moves in the wrong direction.

### 7.3 Recovering Approximate Solutions

Figure 10 shows a run of the algorithm where we fail to get e-convergence to an exact solution. In section 9.4 we will describe one possible strategy, namely *tightening the relaxation*, which can be used to produce an exact solution in these cases. Another obvious strategy, which is approximate, is to simply choose the best primal solution generated after $k$ iterations of the algorithm, for some fixed $k$: i.e., to choose $y^{(k')}, l(y^{(k')})$ where

$$k' = \operatorname*{argmax}_{k' \leq k} f(y^{(k')}) + g(l(y^{(k')}))$$

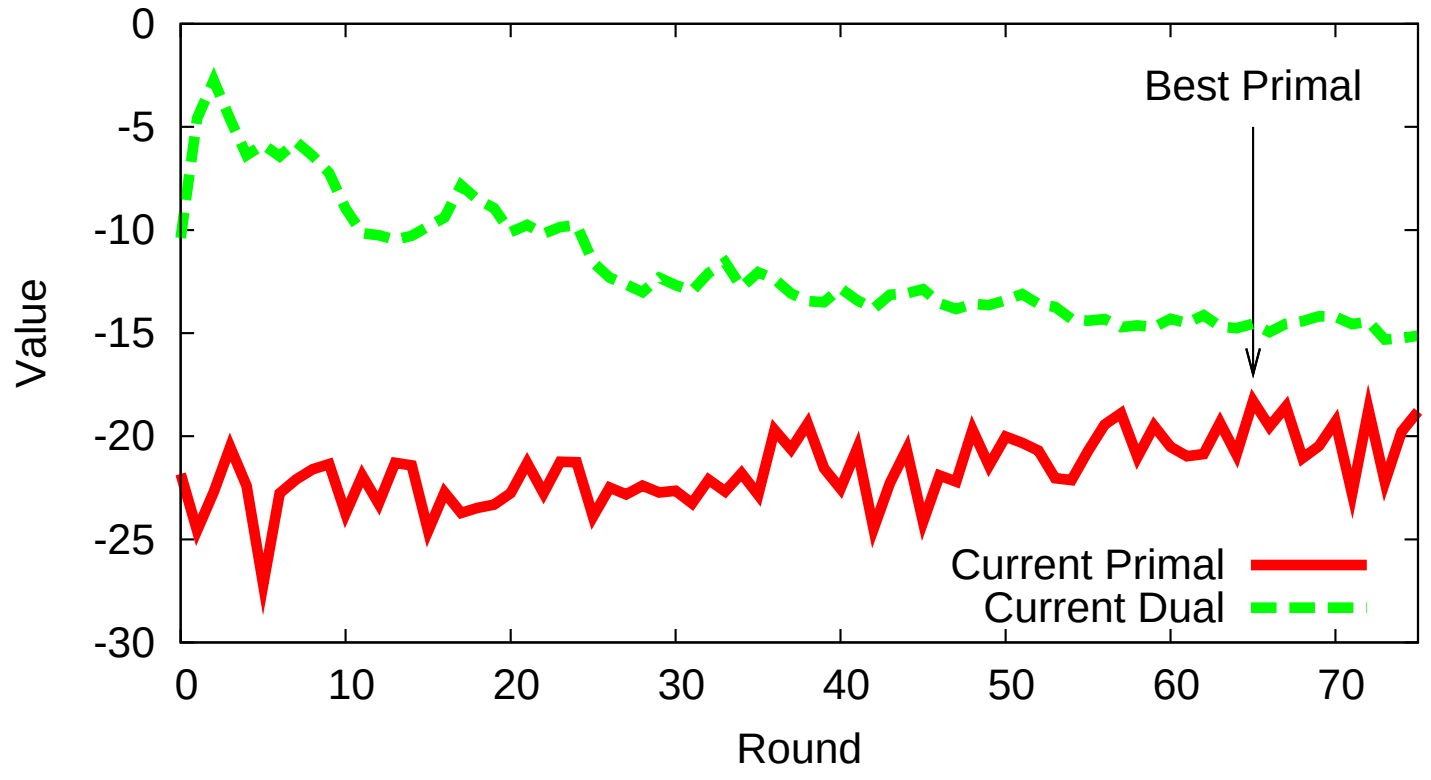

Figure 10: Graph showing the dual value $L(u^{(k)})$ and primal value $f(y^{(k)}) + g(l(y^{(k)}))$, versus iteration number $k$, for the subgradient algorithm on a translation example from the work of Rush and Collins (2011), where the method does not e-converge to an exact solution.

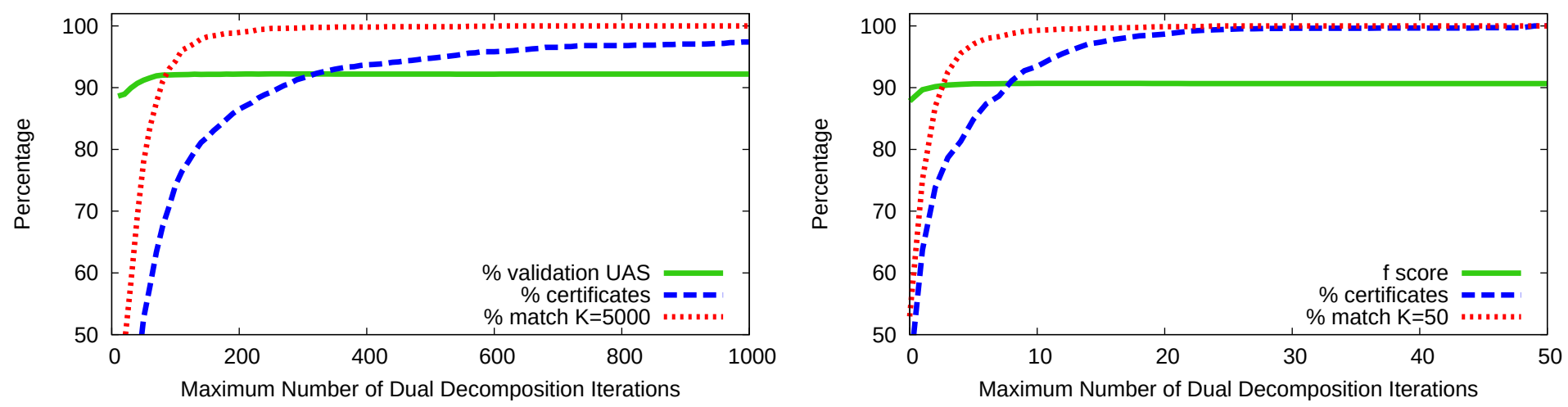

Figure 11: Figures showing effects of early stopping for the non-projective parsing algorithm of Koo et al. (2010) (left graph) and combined constituency and dependency parsing (right graph). In each case we plot three quantities versus the number of iterations, $k$: 1) the accuracy (UAS or f-score); 2) the percentage of cases where the algorithm e-converges giving an exact solution, with a certificate of optimality; 3) the percentage of cases where the best primal solution up to the $k$'th iteration is the same as running the algorithm to e-convergence.

As described before, we can use $L_k^* - p_k^*$ as an upper bound on the difference between this approximate solution and the optimal solution.

### 7.4 Early Stopping

It is interesting to also consider the strategy of returning the best primal solution early in the algorithm in cases where the algorithm *does* eventually e-converge to an exact solution. In practice, this strategy can sometimes produce a high quality solution, albeit without a certificate of optimality, faster than running the algorithm to e-convergence. Figure 11 shows graphs for two problems: non-projective dependency parsing (Koo et al., 2010), and combined constituency and dependency parsing (Rush et al., 2010). In each case we show how three quantities vary with the number of iterations of the algorithm. The first quantity is the percentage of cases where the algorithm e-converges, giving an exact solution, with a certificate of optimality. For combined constituency and dependency parsing it takes roughly 50 iterations for most (over 95%) of cases to e-converge; the second algorithm takes closer to 1000 iterations.

In addition, we show graphs indicating the quality of the best primal solution generated up to iteration $k$ of the algorithm, versus the number of iterations, $k$. An "early stopping" strategy would be to pick some fixed value for $k$, and to simply return the best primal solution generated in the first $k$ iterations of the algorithm. We first plot the accuracy (f-score, or dependency accuracy respectively) for the two models under early stopping: we can see that accuracy very quickly asymptotes to its optimal value, suggesting that returning a primal solution before e-convergence can often yield high quality solutions. We also plot the percentage of cases where the primal solution returned is in fact identical to the primal solution returned when the algorithm is run to e-convergence. We again see that this curve asymptotes quickly, showing that in many cases the early stopping strategy does in fact produce the optimal solution, albeit without a certificate of optimality.

## 8. Alternatives to Subgradient Optimization

This tutorial has focused on subgradient methods for optimization of the dual objective. Several alternative optimization algorithms have been proposed in the machine learning literature; in this section we give an overview of these approaches.

Wainwright, Jaakkola, and Willsky (2005) describe an early and important algorithm for Markov random fields (MRFs) based on LP relaxations, *tree-reweighted message passing* (TRW). Following the work of Kolmogorov (2006), we use TRW-E to refer to the edge-based variant of TRW, and TRW-T to refer to the tree-based algorithm. Kolmogorov (2006) derives a further variant, TRW-S (the "S" refers to the sequential nature of the algorithm). All three algorithms—TRW-E, TRW-T, and TRW-S—are motivated by the LP relaxation for MRFs, but none of them have a guarantee of converging to the optimal value of the LP. TRW-S has the strongest guarantee of the three algorithms, namely that it monotonically improves the dual value, but it may not converge to the optimal dual value.

Yanover et al. (2006) describe experiments comparing TRW-based algorithms to generic LP solvers for MRF problems (specifically, the LP solver they use is CPLEX[13]). The TRW-based algorithms are considerably more efficient than CPLEX, due to the fact that the TRW-based methods leverage the underlying structure of the MRF problem. The various Lagrangian relaxation algorithms described in the current paper can all be viewed as specialized algorithms for solving LP relaxations, which explicitly leverage combinatorial structure within the underlying problem.

Komodakis et al. (2007, 2011) give experiments comparing the subgradient method to the TRW-S and TRW-T algorithms. In these experiments TRW-S generally performs better than TRW-T. In several cases TRW-S finds the optimal dual solution faster than the subgradient method; in other cases TRW-S appears to get stuck (as expected given its lack of convergence guarantee), while the subgradient method finds the global optimum. Overall, the subgradient method is competitive with TRW-S: it may initially make slower progress on the dual objective, but has the benefit of guaranteed convergence to the global optimum of the LP relaxation.

13. http://www.ilog.com/products/cplex/

Another important class of algorithms for optimizing the dual of the LP are block coordinate descent algorithms: for example the MPLP algorithm of Globerson and Jaakkola (2007). See the work of Sontag et al. (2010) for a discussion of these methods. Like TRW-S, the MPLP algorithm is guaranteed to monotonically improve the dual value, but is not guaranteed to converge to the global optimum of the MRF LP. In several experimental settings, the MPLP algorithm produces better dual values in early iterations than subgradient methods, but can get stuck at a non-optimal solution (Jojic, Gould, & Koller, 2010; Martins, Figueiredo, Aguiar, Smith, & Xing, 2011; Meshi & Globerson, 2011). Another complication of MPLP is that it requires computing max-marginals for the sub-problems at each iteration instead of MAP assignments. Max-marginals may be slower to compute in practice, and for some combinatorial problems computation may be asymptotically slower. (For example, for the directed spanning tree models from Koo et al., 2010, the MAP problem can be solved in $O(n^2)$ time where $n$ is the length of the input sentence, but we are not aware of an algorithm that solves the max-marginal problem in better than $O(n^4)$ time.)

In other work, Jojic et al. (2010) describe an accelerated method for MRF inference, using the method of Nesterov (2005) to smooth the objective in the underlying decomposition. The method has a relatively fast rate of convergence ($O(1/\epsilon)$ time to reach a solution that is $\epsilon$-close to optimal). Experiments from the work of Jojic et al. (2010) show a decrease in the number of iterations required compared to subgradient; however in the work of Martins et al. (2011) the accelerated method requires more iterations than the subgradient algorithm. In both sets of experiments, MPLP makes more initial progress than either method. Accelerated subgradient also requires computing sub-problem marginals, which has similar disadvantages as MPLP's requirement of max-marginals.

Recently, Martins et al. (2011) proposed an augmented Lagrangian method for inference using the alternating direction method of multipliers (ADMM). See the tutorial of Boyd, Parikh, Chu, Peleato, and Eckstein (2011) on ADMM. The augmented Lagrangian method further extends the objective with a quadratic penalty term representing the amount of constraint violation. ADMM is a method for optimizing this augmented problem that is able to maintain similar decomposibility properties as dual decomposition. Like the subgradient method, ADMM is guaranteed to find the optimum of the LP relaxation. Martins et al. (2011) show empirically that ADMM requires a comparable number of iterations to MPLP to find a good primal solution, while still being guaranteed to optimize the LP. A challenge of ADMM is that the extra quadratic term may complicate sub-problem decoding, for example it is not clear how to directly decode the parsing problems presented in this work with a quadratic term in the objective. Several alternative approaches have been proposed: Martins et al. (2011) binarize the combinatorial sub-problems into binary-valued factor graphs; Meshi and Globerson (2011) avoid the problem by instead applying ADMM to the dual of the LP; Martins (2012) and Das et al. (2012) use an iterative active set method that utilizes MAP solutions of the original sub-problems to solve the quadratic version. Martins (2012) also describes recent results on ADMM that give a $O(1/\epsilon)$ bound for relaxed primal convergence.

## 9. The Relationship to Linear Programming Relaxations

This section describes the close relationship between the dual decomposition algorithm and linear programming relaxations. This connection will be very useful in understanding the behavior of the algorithm, and in particular in understanding the cases where the algorithm does not e-converge to an exact solution. In addition, it will suggest strategies for "tightening" the algorithm until an exact solution is found.

### 9.1 The Linear Programming Relaxation

We continue to use the algorithm from section 4 as an example; the generalization to other problems is straightforward. First, define the set

$$\Delta_y = \{\alpha : \alpha \in \mathbb{R}^{|\mathcal{Y}|}, \sum_y \alpha_y = 1, \forall y \; 0 \leq \alpha_y \leq 1\}$$

Thus $\Delta_y$ is a simplex, corresponding to the set of probability distributions over the finite set $\mathcal{Y}$. Similarly, define

$$\Delta_z = \{\beta : \beta \in \mathbb{R}^{|\mathcal{Z}|}, \sum_z \beta_z = 1, \forall z \; 0 \leq \beta_z \leq 1\}$$

as the set of distributions over the set $\mathcal{Z}$.

We now define a new optimization problem, as follows:

**Optimization Problem 5** *Find*

$$\max_{\alpha \in \Delta_y, \beta \in \Delta_z} \sum_y \alpha_y f(y) + \sum_z \beta_z g(z) \tag{29}$$

*such that for all $i, t$,*

$$\sum_y \alpha_y y(i,t) = \sum_z \beta_z z(i,t) \tag{30}$$

This optimization problem is a linear program: the objective in Eq. 29 is linear in the variables $\alpha$ and $\beta$; the constraints in Eq. 30, together with the constraints in the definitions of $\Delta_y$ and $\Delta_z$, are also linear in these variables.

This optimization problem is very similar to our original problem, optimization problem 1. To see this, define $\Delta'_y$ as follows:

$$\Delta'_y = \{\alpha : \alpha \in \mathbb{R}^{|\mathcal{Y}|}, \sum_y \alpha_y = 1, \forall y \; \alpha_y \in \{0,1\}\}$$

Thus $\Delta'_y$ is a subset of $\Delta_y$, where the constraints $0 \leq \alpha_y \leq 1$ have been replaced by $\alpha_y \in \{0,1\}$. Define $\Delta'_z$ similarly. Consider the following optimization problem, where we replace $\Delta_y$ and $\Delta_z$ in Eq. 29 by $\Delta'_y$ and $\Delta'_z$ respectively:

**Optimization Problem 6** *Find*

$$\max_{\alpha \in \Delta'_y, \beta \in \Delta'_z} \sum_y \alpha_y f(y) + \sum_z \beta_z g(z) \tag{31}$$

*such that for all $i, t$,*

$$\sum_y \alpha_y y(i,t) = \sum_z \beta_z z(i,t) \tag{32}$$

This new problem is equivalent to our original problem, optimization problem 1: choosing vectors $\alpha \in \Delta'_y$ and $\beta \in \Delta'_z$ is equivalent to choosing a single parse in $\mathcal{Y}$, and a single POS sequence in $z$. In this sense, optimization problem 5 is a relaxation of our original problem, where constraints of the form $\alpha_y \in \{0, 1\}$ and $\beta_z \in \{0, 1\}$ are replaced with constraints of the form $0 \leq \alpha_y \leq 1$ and $0 \leq \beta_z \leq 1$.

Note that optimization problem 6 is an *integer linear program*, because the objective is again linear in the $\alpha$ and $\beta$ variables, and the constraints on these variables combine linear constraints with integer constraints (that each $\alpha_y$ and $\beta_z$ must be either 0 or 1). It is also worth noting that $\Delta_y$ is actually the *convex hull* of the finite set $\Delta'_y$. The points in $\Delta'_y$ form the vertices of the polytope $\Delta_y$.

A useful theorem, which is central to the relationship between linear programming and combinatorial optimization problems, is the following:

**Theorem 6** *For any finite set* $\mathcal{Y}$*, and any function* $f : \mathcal{Y} \rightarrow \mathbb{R}$*,*

$$\max_{y \in \mathcal{Y}} f(y) = \max_{\alpha \in \Delta_y} \sum_{y \in \mathcal{Y}} \alpha_y f(y)$$

*where* $\Delta_y$ *is as defined above.*

The proof is simple, and is given in Appendix A.4.

### 9.2 The Dual of the New Optimization Problem

We now describe the dual problem for the linear program in Eqs. 29 and 30. This will again be a function $M(u)$ of a vector of dual variables $u = \{u(i, t) : i \in \{1 \ldots n\}, t \in \mathcal{T}\}$. A crucial result will be that *the two dual functions* $M(u)$ *and* $L(u)$ *are identical.*

Our new Lagrangian is

$$\begin{aligned} M(u, \alpha, \beta) &= \sum_y \alpha_y f(y) + \sum_z \beta_z g(z) + \sum_{i,t} u(i, t) \left( \sum_y \alpha_y y(i, t) - \sum_z \beta_z z(i, t) \right) \\ &= \left( \sum_y \alpha_y f(y) + \sum_{i,t} u(i, t) \sum_y \alpha_y y(i, t) \right) \\ &\quad + \left( \sum_z \beta_z g(z) - \sum_{i,t} u(i, t) \sum_z \beta_z z(i, t) \right) \end{aligned}$$

The new dual objective is

$$M(u) = \max_{\alpha \in \Delta_y, \beta \in \Delta_z} M(u, \alpha, \beta)$$

Note that once again we have simply maximized out over the primal ($\alpha$ and $\beta$) variables, ignoring the constraints in Eq. 30. The dual problem is to find

$$\min_u M(u)$$

Two theorems regarding the dual problem are then as follows:

**Theorem 7** *Define* $(\alpha^*, \beta^*)$ *to be the solution to the optimization problem in Eqs. 29 and 30. Then*

$$\min_u M(u) = \sum_y \alpha^*_y f(y) + \sum_z \beta^*_z g(z)$$

*Proof.* This follows immediately by results from linear programming duality see the textbook of Korte and Vygen (2008) for more details. □

Note that we now have *equality* in the above, in contrast to our previous result,

$$\min_u L(u) \geq f(y^*) + g(z^*)$$

where the dual function only gave an upper bound on the best primal solution.

Our second theorem is as follows:

**Theorem 8** *For any value of* $u$*,*

$$M(u) = L(u)$$

Thus the two dual functions are identical. Given that the subgradient algorithm we have described minimizes $L(u)$, it therefore also minimizes the dual of the linear program in Eqs. 29 and 30.

*Proof.* We have

$$\begin{aligned}
M(u) &= \max_{\alpha \in \Delta_y} \left( \sum_y \alpha_y f(y) + \sum_{i,t} u(i,t) \sum_y \alpha_y y(i,t) \right) + \\
&\quad \max_{\beta \in \Delta_z} \left( \sum_z \beta_z g(z) - \sum_{i,t} u(i,t) \sum_z \beta_z z(i,t) \right) \\
&= \max_{y \in \mathcal{Y}} \left( f(y) + \sum_{i,t} u(i,t) y(i,t) \right) + \\
&\quad \max_{z \in \mathcal{Z}} \left( g(z) - \sum_{i,t} u(i,t) z(i,t) \right) \\
&= L(u)
\end{aligned}$$

where we have used theorem 6 to give

$$\max_{\alpha \in \Delta_y} \left( \sum_y \alpha_y f(y) + \sum_{i,t} u(i,t) \sum_y \alpha_y y(i,t) \right) = \max_{y \in \mathcal{Y}} \left( f(y) + \sum_{i,t} u(i,t) y(i,t) \right)$$

and we have used a similar result to replace the $\max$ over $\Delta_z$ by the $\max$ over $\mathcal{Z}$. □

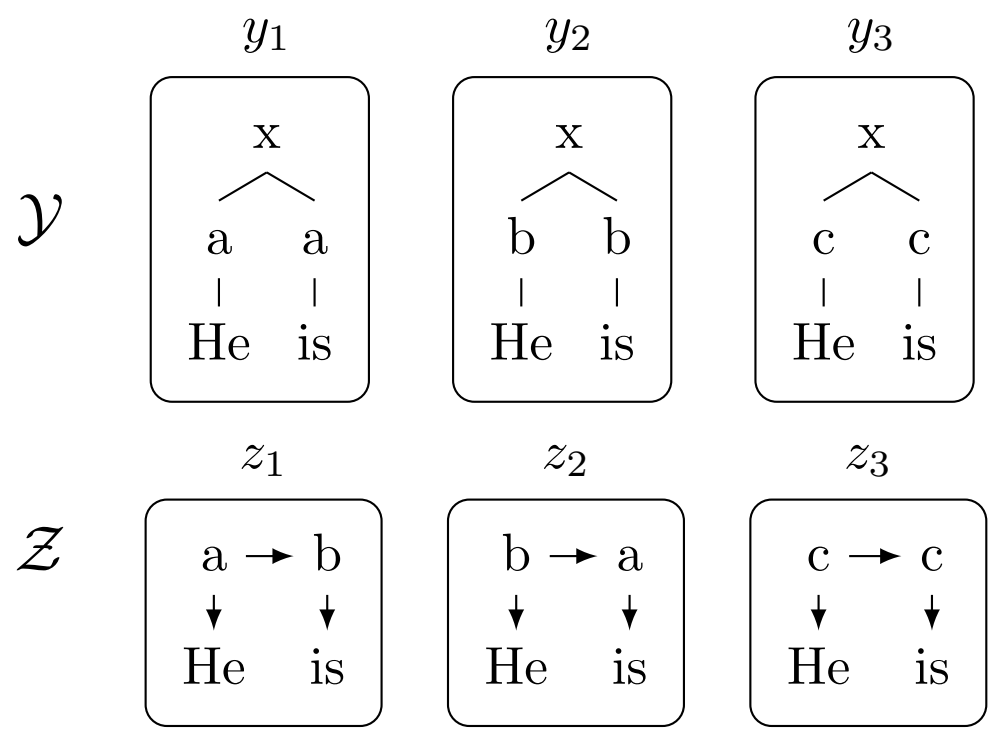

Figure 12: A simple example with three possible parse trees and three possible POS sequences.

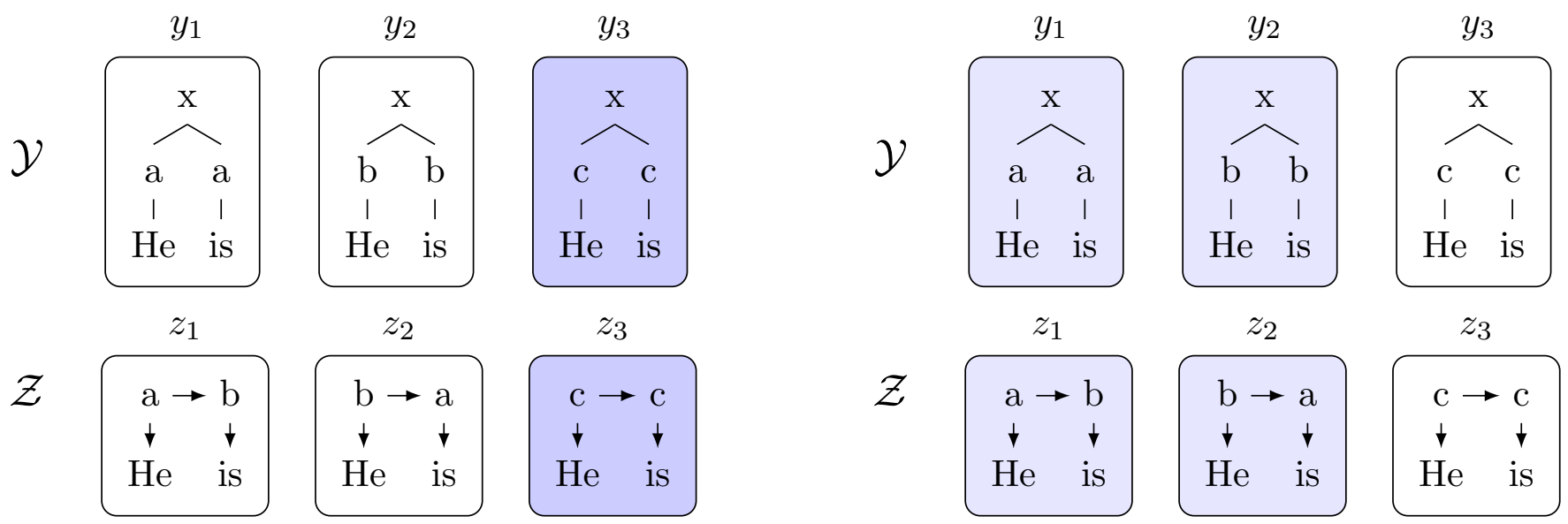

Figure 13: Illustration of two solutions that satisfy the constraints in Eq. 30. On the left, the solution $\alpha = [0, 0, 1]$, $\beta = [0, 0, 1]$ puts weight 1 on $y_3$ and $z_3$. On the right, the fractional solution $\alpha = [0.5, 0.5, 0]$ and $\beta = [0.5, 0.5, 0]$ puts 0.5 weight on $y_1/y_2$ and $z_1/z_2$.

### 9.3 An Example

We now give an example that illustrates these ideas. Through this example we will also illustrate what happens when the algorithm fails to e-converge.

We assume that there are three possible parse trees, $\mathcal{Y} = \{y_1, y_2, y_3\}$, and three possible tag sequences, $\mathcal{Z} = \{z_1, z_2, z_3\}$, shown in Figure 12. We will write distributions over these sets as vectors such as $\alpha = [0, 0, 1]$ or $\beta = [0.5, 0.5, 0]$.

Now consider pairs of vectors $(\alpha, \beta)$ that satisfy the constraints in Eq. 30. Figure 13 illustrates two possible solutions. One such pair, which we will denote as $(\alpha^1, \beta^1)$, is $\alpha^1 = [0, 0, 1]$, $\beta^1 = [0, 0, 1]$. It is easily verified that under this definition

$$\sum_{y\in\mathcal{Y}} \alpha^1_y y(1, c) = \sum_{z\in\mathcal{Y}} \beta^1_z z(1, c) = \sum_{y\in\mathcal{Y}} \alpha^1_y y(2, c) = \sum_{z\in\mathcal{Y}} \beta^1_z z(2, c) = 1$$

with all other expected values being equal to 0: hence $(\alpha^1, \beta^1)$ satisfies the constraints. This potential solution is *integral*, in that it puts weight 1 on a single parse tree/POS-tag sequence, with all other structures having weight 0.

A second pair that satisfies the constraints is $\alpha^2 = [0.5, 0.5, 0]$, $\beta^2 = [0.5, 0.5, 0]$. Under these definitions,

$$\sum_{y\in\mathcal{Y}} \alpha^2_y y(1, a) = \sum_{z\in\mathcal{Y}} \beta^2_z z(1, a) = \sum_{y\in\mathcal{Y}} \alpha^2_y y(1, b) = \sum_{z\in\mathcal{Y}} \beta^2_z z(1, b) = 0.5$$

and

$$\sum_{y\in\mathcal{Y}} \alpha^2_y y(2, a) = \sum_{z\in\mathcal{Y}} \beta^2_z z(2, a) = \sum_{y\in\mathcal{Y}} \alpha^2_y y(2, b) = \sum_{z\in\mathcal{Y}} \beta^2_z z(2, b) = 0.5$$

with all other expected values being equal to 0. The pair $(\alpha^2, \beta^2)$ is a *fractional solution*, in that it puts fractional (0.5) weight on some structures.

Next, consider different definitions for the functions $f(y)$ and $g(z)$. Consider first the definitions $f = [0, 0, 1]$ and $g = [0, 0, 1]$ (we write $f = [0, 0, 1]$ as shorthand for $f(y_1) = 0$, $f(y_2) = 0$, $f(y_3) = 1$). The solution to the problem in Eqs. 29 and 30 is then the pair $(\alpha^1, \beta^1)$.

Alternatively, consider the definitions $f = [1, 1, 2]$ and $g = [1, 1, -2]$. In this case the following situation arises:

- The pair $(\alpha^1, \beta^1)$ achieves score 0 under the objective in Eq. 29, whereas the pair $(\alpha^2, \beta^2)$ achieves a score of 2. *Thus the solution to the problem in Eqs. 29 and 30 is $(\alpha^2, \beta^2)$, which is a fractional solution.*

- By theorem 7, $\min_u M(u)$ is equal to the value for the optimal primal solution, i.e., $\min_u M(u) = 2$. Hence $\min_u L(u) = 2$.

- In contrast, the solution to the original optimization problem 1 is $(y^*, z^*) = (y_3, z_3)$: in fact, $(y_3, z_3)$ is the only pair of structures that satisfies the constraints $y(i, t) = z(i, t)$ for all $(i, t)$. Thus $f(y^*) + g(z^*) = 0$. We have

  $$\min_u L(u) = 2 > f(y^*) + g(z^*) = 0$$

  Thus there is a clear gap between the minimum dual value, and the score for the optimal primal solution.

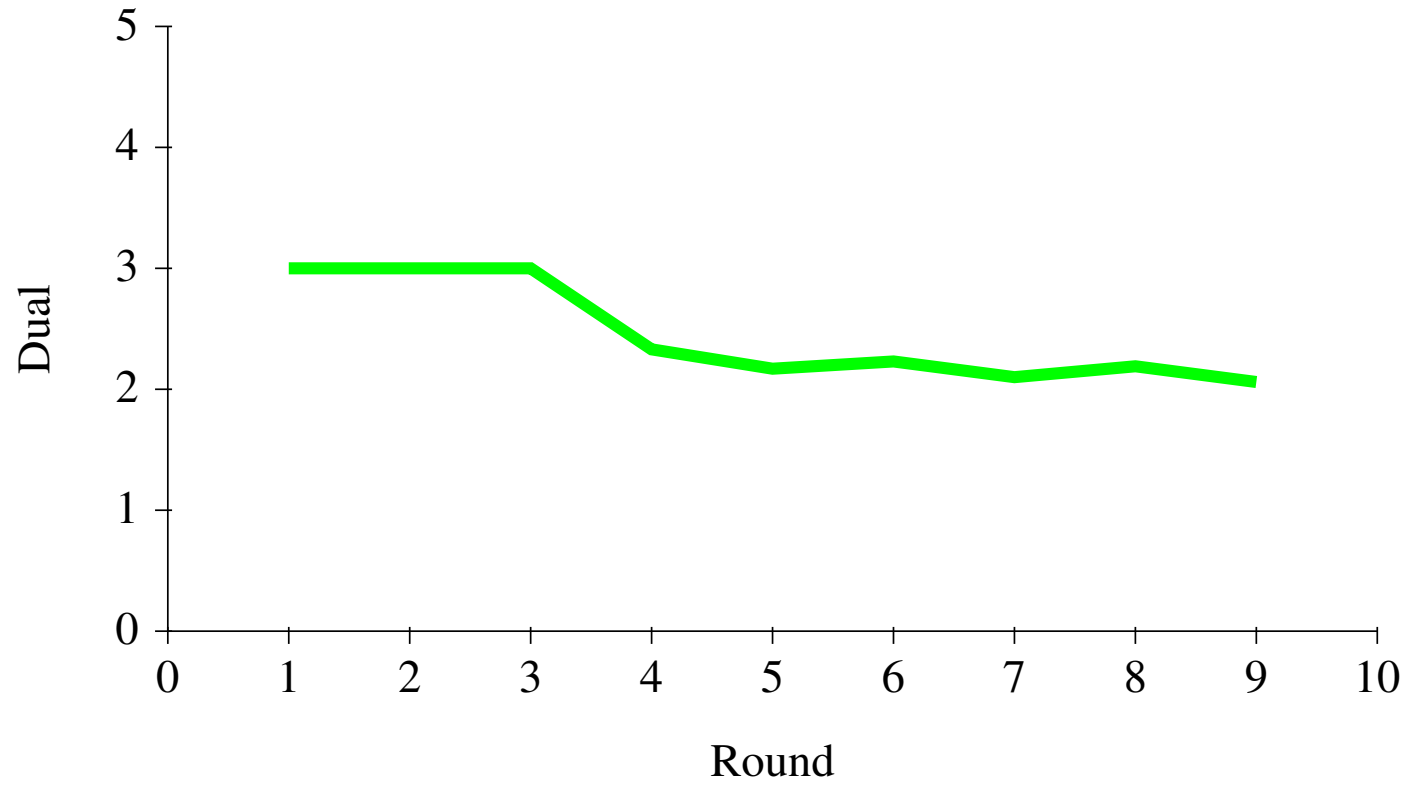

| $k$ | $y^{(k)}$ | $z^{(k)}$ |
|---|---|---|
| 1 | $y_3$ | $z_2$ |
| 2 | $y_2$ | $z_1$ |
| 3 | $y_1$ | $z_1$ |
| 4 | $y_1$ | $z_1$ |
| 5 | $y_2$ | $z_2$ |
| 6 | $y_1$ | $z_1$ |
| 7 | $y_2$ | $z_2$ |
| 8 | $y_1$ | $z_1$ |
| 9 | $y_2$ | $z_2$ |

Figure 14: Figures showing progress of the subgradient algorithm for $f = [1, 1, 2]$ and $g = [1, 1, -2]$. The graph shows the dual value $L(u^{(k)})$ versus number of iterations $k$. The table shows the hypotheses $y^{(k)}$ and $z^{(k)}$ versus number of iterations $k$. In later iterations the method alternates between hypotheses $(y_1, z_1)$ and $(y_2, z_2)$.

Figure 14 shows a trace of the subgradient method on this problem. After nine iterations the method has reached $L(u^{(9)}) = 2.06$, which is close to the optimal dual value. In this case, however, the algorithm does not reach agreement between the structures $y^{(k)}$ and $z^{(k)}$. Instead, it reaches a point where it alternates between solutions $(y^{(1)}, z^{(1)})$, and $(y^{(2)}, z^{(2)})$. Thus the dual d-converges to its minimum value, but the primal solutions generated alternate between the structures $y_1, y_2, z_1, z_2$ that have greater than 0 weight in the fractional solution $(\alpha^2, \beta^2)$. This behavior is typical of cases where there is a duality gap, i.e., where $\min_u L(u)$ is strictly greater than $f(y^*) + g(z^*)$.

### 9.4 Fixing E-Convergence: Tightening Approaches

We now describe a "tightening" approach that can be used to fix the issue of non-convergence given in the previous example.

Consider again the problem of integrated CFG parsing and HMM tagging. Assume that the input sentence is of length $n$. The first approach is as follows. We introduce new variables $y(i, t_1, t_2)$ for $i = 1 \dots (n-1)$, $t_1 \in \mathcal{T}$, $t_2 \in \mathcal{T}$, with $y(i, t_1, t_2) = 1$ if $y(i, t_1) = 1$ and $y(i+1, t_2) = 1$, 0 otherwise. Thus the new variables track tag *bigrams*. Similarly, we introduce variables $z(i, t_1, t_2)$ for tag sequences $z \in \mathcal{Z}$. We now define the set of constraints to be

$$y(i, t) = z(i, t)$$

for all $i \in \{1 \dots n\}$, $t \in \mathcal{T}$ (the same constraints as before), and in addition

$$y(i, t_1, t_2) = z(i, t_1, t_2)$$

for all $i \in \{1 \dots n-1\}$, $t_1 \in \mathcal{T}$, $t_2 \in \mathcal{T}$.

We then proceed as before, using Lagrange multipliers $u(i, t)$ to enforce the first set of constraints, and Lagrange multipliers $v(i, t_1, t_2)$ to enforce the second set of constraints. The dual decomposition algorithm will require us to find

$$y^{(k)} = \operatorname*{argmax}_{y \in \mathcal{Y}} f(y) + \sum_{i,t} u^{(k-1)}(i,t) y(i,t) + \sum_{i,t_1,t_2} v^{(k-1)}(i,t_1,t_2) y(i,t_1,t_2) \tag{33}$$

and

$$z^{(k)} = \operatorname*{argmax}_{z \in \mathcal{Z}} g(z) - \sum_{i,t} u^{(k-1)}(i,t) z(i,t) - \sum_{i,t_1,t_2} v^{(k-1)}(i,t_1,t_2) z(i,t_1,t_2) \tag{34}$$

at each iteration, followed by updates of the form

$$u^{(k)}(i,t) \leftarrow u^{(k-1)}(i,t) - \delta(y^{(k)}(i,t) - z^{(k)}(i,t))$$

and

$$v^{(k)}(i,t_1,t_2) \leftarrow v^{(k-1)}(i,t_1,t_2) - \delta(y^{(k)}(i,t_1,t_2) - z^{(k)}(i,t_1,t_2))$$

It can be shown that if $g(z)$ is defined through a bigram HMM model, the above method is guaranteed to e-converge to an exact solution. In fact, the underlying LP relaxation is now *tight*, in that only integral solutions are possible.

The problem with this approach is that finding the $\operatorname{argmax}$ in Eq. 33 is now expensive, due to the $v(i,t_1,t_2)y(i,t_1,t_2)$ terms: in fact, it requires the exact dynamic programming algorithm for intersection of a bigram HMM with a PCFG. Thus we end up with an algorithm that is at least as expensive as integration of a bigram HMM with a PCFG using the construction of Bar-Hillel et al. (1964).[14]

A second approach, which may be more efficient, is as follows. Rather than introducing all constraints of the form of Eq. 33, we might introduce a few selected constraints. As an example, with the previous non-convergent example we might add the single constraint

$$y(1,a,b) = z(1,a,b)$$

We have a single Lagrange multiplier $v(1,a,b)$ for this new constraint, and the dual decomposition algorithm requires the following steps at each iteration:

$$y^{(k)} = \operatorname*{argmax}_{y \in \mathcal{Y}} f(y) + \sum_{i,t} u^{(k-1)}(i,t) y(i,t) + v^{(k-1)}(1,a,b) y(1,a,b) \tag{35}$$

and

$$z^{(k)} = \operatorname*{argmax}_{z \in \mathcal{Z}} g(z) - \sum_{i,t} u^{(k-1)}(i,t) z(i,t) - v^{(k-1)}(1,a,b) z(1,a,b) \tag{36}$$

and updates

$$u^{(k)}(i,t) \leftarrow u^{(k-1)}(i,t) - \delta(y^{(k)}(i,t) - z^{(k)}(i,t))$$

and

$$v^{(k)}(1,a,b) \leftarrow v^{(k-1)}(1,a,b) - \delta(y^{(k)}(1,a,b) - z^{(k)}(1,a,b))$$

Figure 15 shows a run of the subgradient algorithm with this single constraint added. The fractional solution $(\alpha^2, \beta^2)$ is now eliminated, and the method e-converges to the correct solution.

Two natural questions arise:

14. If $g(z)$ is defined through a bigram HMM, then clearly nothing has been gained in efficiency over the Bar-Hillel et al. (1964) method. If $g(z)$ is more complex, for example consisting of a trigram model, the dual decomposition method may still be preferable.

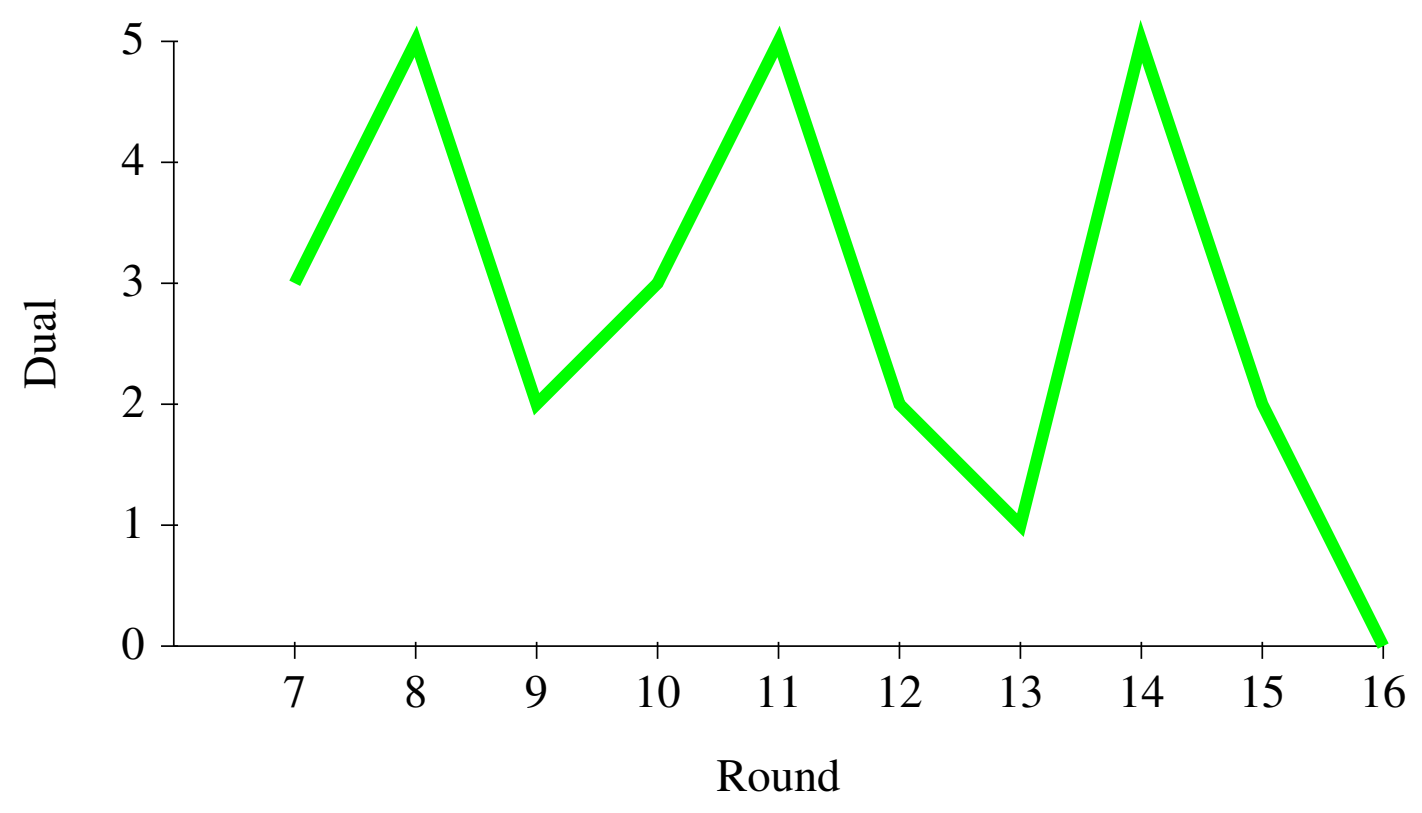

| $k$ | $y^{(k)}$ | $z^{(k)}$ |
|---|---|---|
| 7 | $y_3$ | $z_2$ |
| 8 | $y_2$ | $z_3$ |
| 9 | $y_1$ | $z_2$ |
| 10 | $y_3$ | $z_1$ |
| 11 | $y_2$ | $z_3$ |
| 12 | $y_1$ | $z_2$ |
| 13 | $y_3$ | $z_1$ |
| 14 | $y_2$ | $z_3$ |
| 15 | $y_1$ | $z_2$ |
| 16 | $y_3$ | $z_3$ |

Figure 15: Figures showing progress of the subgradient algorithm for $f(y) = [1, 1, 2]$ and $g(z) = [1, 1, -2]$, with the additional constraint $y(1, a, b) = z(1, a, b)$ incorporated in the Lagrangian. The graph shows the dual value $L(u^{(k)})$ versus number of iterations $k$. The table shows the hypotheses $y^{(k)}$ and $z^{(k)}$ versus number of iterations $k$.

- *Which constraints should be added?* One strategy is to first run the subgradient method with the basic constraints, as shown in Figure 14. Some heuristic is used to determine that the dual is no longer decreasing at a significant rate. At that point, it can be determined that the algorithm is oscillating between solutions $(y_1, z_1)$ and $(y_2, z_2)$, and that the additional constraint $y(1, a, b) = z(1, a, b)$ would rule out these solutions; hence this constraint is added.

- *When is this more efficient than adding all constraints in Eq. 33?* Our toy example is too simple to illustrate the benefit of only adding selected constraints. To understand the benefit, consider the case where the sentence length $n$ is reasonably large. In that case, we might add bigram constraints at only a few positions in the sentence: in practice the CKY decoding algorithm will only need to introduce the Bar-Hillel et al. (1964) machinery at these selected points, which can be much more efficient that introducing all constraints.

For examples of methods that tighten dual decomposition/Lagrangian relaxation techniques using additional constraints, see the work of Sontag, Meltzer, Globerson, Jaakkola, and Weiss (2008), Rush and Collins (2011), Chang and Collins (2011), and Das et al. (2012). This is related to previous work on non-projective dependency parsing (Riedel & Clarke, 2006) that incrementally adds constraints to an integer linear program solver.

### 9.5 Compact Linear Programs

The LP relaxations that we have described have a very large set of variables: that is, one variable for each member of $\mathcal{Y}$ and $\mathcal{Z}$. In most cases of interest, the sets $\mathcal{Y}$ and $\mathcal{Z}$ will be exponential in size.

In this section we describe how to derive equivalent linear programs with far fewer variables. This is a problem of practical interest: for many problems, we have found it beneficial to implement the underlying LP relaxation within a generic LP solver, as a way of debugging dual decomposition algorithms. This is practical with the compact LPs that we describe in this section, but is clearly impractical with the exponential-size linear programs described in the previous section.

First, consider the abstract description of Lagrangian relaxation given in section 3. The LP relaxation was

$$\operatorname*{argmax}_{\mu \in \mathcal{Q}} \mu \cdot \theta$$

where

$$\mathcal{Q} = \{y : y \in \text{Conv}(\mathcal{Y}') \text{ and } Ay = b\}$$

where $A \in \mathbb{R}^{p \times d}$ and $b \in \mathbb{R}^p$. Recall that $\text{Conv}(\mathcal{Y}')$ is the convex hull of the set $\mathcal{Y}'$. Next, assume that $\text{Conv}(\mathcal{Y}')$ can itself be defined through a polynomial number of linear constraints: that is,

$$\text{Conv}(\mathcal{Y}') = \{y \in \mathbb{R}^d : Cy = e, y \geq 0\} \tag{37}$$

for some $C \in \mathbb{R}^{q \times d}$ and $e \in \mathbb{R}^q$, where the number of constraints, $q$, is polynomial. In this case we have an explicit characterization of the set $\mathcal{Q}$ as

$$\mathcal{Q} = \{y \in \mathbb{R}^d : Cy = e, y \geq 0 \text{ and } Ay = b\}$$

Because $d$, $p$, and $q$ are all polynomial in size, the resulting linear program is polynomial in size. In this sense it is "compact".

The remaining question is whether a characterization of the form of Eq. 37 exists, and if so, how it is defined. Recall that we made the assumption that for any value of $\theta$,

$$\operatorname*{argmax}_{y \in \mathcal{Y}'} y \cdot \theta \tag{38}$$

can be found using a combinatorial algorithm. For many combinatorial algorithms, there are LP formulations that are polynomial in size: these formulations lead directly to definitions of $C$ and $e$.[15] For example, Martin, Rardin, and Campbell (1990) give such a construction for dynamic programming algorithms, which includes parsing algorithms for weighted context-free grammars, the Viterbi algorithm, and other dynamic programs used in NLP. Martins et al. (2009) make use of a construction for directed spanning trees (see also Magnanti & Wolsey, 1994), and apply it to non-projective dependency parsing. Korte and Vygen (2008) describe many other such constructions. In short, given a combinatorial algorithm that solves the problem in Eq. 38, it is often straightforward to find a recipe for constructing a pair $(C, e)$ that completely characterizes $\text{Conv}(\mathcal{Y}')$.

It is straightforward to extend this idea to the LP for dual decomposition. Consider again our running example, (optimization problem 1),

$$\operatorname*{argmax}_{y \in \mathcal{Y}, z \in \mathcal{Z}} f(y) + g(z)$$

such that for all $i = 1 \ldots n, t \in \mathcal{T}$,

$$y(i, t) = z(i, t)$$

Rush et al. (2010) give a full description of the compact LP for this problem: we give a sketch here.

15. There is one subtlety here: in some cases additional auxilliary variables may need to be introduced. See for example the spanning tree construction of Magnanti and Wolsey (1994). However the number of auxilliary variables is generally polynomial in number, hence this is benign.

Define each $y$ to be a vector in $\{0,1\}^d$ that specifies which context-free rules $y$ contains. It follows that $\mathcal{Y}$ is a subset of $\{0,1\}^d$. We then have

$$f(y) = y \cdot \theta$$

where $\theta \in \mathbb{R}^d$ is a vector specifying the weight for each rule. Similarly, define $z$ to be a vector in $\{0,1\}^{d'}$ that specifies the trigrams that $z$ contains (assuming that $g(z)$ is a trigram tagging model). It follows that $\mathcal{Z}$ is a subset of $\{0,1\}^{d'}$. We can then write

$$g(z) = z \cdot \theta'$$

for some $\theta' \in \mathbb{R}^{d'}$. The compact LP is then

$$\operatorname*{argmax}_{\mu \in \text{Conv}(\mathcal{Y}), \nu \in \text{Conv}(\mathcal{Z})} \quad \mu \cdot \theta + \nu \cdot \theta'$$

such that for all $i = 1 \ldots n, t \in \mathcal{T}$,

$$\mu(i,t) = \nu(i,t)$$

Again, the existence of combinatorial algorithms for the problems $\text{argmax}_{y \in \mathcal{Y}}\, y \cdot \theta$ and $\text{argmax}_{z \in \mathcal{Z}}\, z \cdot \theta'$ implies explicit representations

$$\text{Conv}(\mathcal{Y}) = \{\mu \in \mathbb{R}^d : A\mu = b, \mu \geq 0\}$$

and

$$\text{Conv}(\mathcal{Z}) = \{\nu \in \mathbb{R}^{d'} : C\nu = e, \nu \geq 0\}$$

where $A, b, C$ and $e$ are polynomial in size. Rush et al. (2010) describe this construction in detail for the case where a weighted CFG is combined with a finite-state tagger.

### 9.6 Summary

To summarize, the key points of this section were as follows:

- We introduced a linear programming problem that was a relaxation of our original problem. The function $L(u)$ was shown to be the dual of this linear programming relaxation.

- In cases where the optimal solution to the underlying LP is fractional, the subgradient method will still d-converge to $\min_u L(u)$. However the primal solutions $(y^{(k)}, z^{(k)})$ will alternate between different solutions that do not satisfy the $y(i,t) = z(i,t)$ constraints.

- In practice, *tightening methods* can be used to improve convergence. These methods selectively introduce constraints in an effort to improve convergence of the method, with the cost of increased complexity in finding $y^{(k)}$ and/or $z^{(k)}$. The precise constraints to be added can be chosen by identifying constraints that are frequently violated during the subgradient method.

- Finally, we described methods that construct a compact linear program that is equivalent to the original LP relaxation. This linear program is often small enough to be solved by a generic LP solver; this can be useful in debugging dual decomposition or Lagrangian relaxation algorithms.

## 10. Conclusions

A broad class of inference problems in statistical NLP and other areas of machine learning are amenable to Lagrangian relaxation (LR) methods. LR methods make use of combinatorial algorithms in combination with linear constraints that are introduced using Lagrange multipliers: iterative methods are used to minimize the resulting dual objective. LR algorithms are simple and efficient, typically involving repeated applications of the underlying combinatorial algorithm, in conjunction with simple additive updates to the Lagrange multipliers. They have well-understood formal properties: the dual objective is an upper bound on the score for the optimal primal solution; there are close connections to linear programming relaxations; and crucially, they have the potential of producing an exact solution to the original inference problem, with a certificate of optimality. Experiments on several NLP problems have shown the effectiveness of LR algorithms for inference: LR methods are often considerably more efficient than existing exact methods, and have stronger formal guarantees than the approximate search methods that are often used in practice.

## Acknowledgments

We thank the anonymous reviewers for helpful comments. Tommi Jaakkola and David Sontag introduced us to dual decomposition and Lagrangian relaxation for inference in probabilistic models; this work would not have happened without them. We have benefited from many discussions with Yin-Wen Chang, Terry Koo, and Roi Reichart, who with Tommi and David were collaborators on our work on dual decomposition/Lagrangian relaxation for NLP. We also thank Shay Cohen, Yoav Goldberg, Mark Johnson, Andre Martins, Ryan McDonald, and Slav Petrov for feedback on earlier drafts of this paper. Columbia University gratefully acknowledges the support of the Defense Advanced Research Projects Agency (DARPA) Machine Reading Program under Air Force Research Laboratory (AFRL) prime contract no. FA8750-09-C-0181. Alexander Rush was supported by a National Science Foundation Graduate Research Fellowship.

## Appendix A. Proofs

In this section we derive various results for the combined parsing and tagging problem. Recall that in this case the Lagrangian is defined as

$$L(u, y, z) = f(y) + g(z) + \sum_{i \in \{1 \ldots n\}, t \in \mathcal{T}} u(i, t)(y(i, t) - z(i, t))$$

and that the dual objective is $L(u) = \max_{y \in \mathcal{Y}, z \in \mathcal{Z}} L(u, y, z)$. Here $n$ is the number of words in the sentence, and $\mathcal{T}$ is a finite set of part-of-speech tags.

We first prove that $L(u)$ is a convex function; we then derive the expression for subgradients of $L(u)$; we then give a convergence theorem for the algorithm in Figure 2, which is a subgradient algorithm for minimization of $L(u)$.

Finally, we give a proof of theorem 6.

### A.1 Proof of Convexity of $L(u)$

The theorem is as follows:

**Theorem 9** *$L(u)$ is convex. That is, for any $u^{(1)} \in \mathbb{R}^d$, $u^{(2)} \in \mathbb{R}^d$, $\lambda \in [0,1]$,*

$$L(\lambda u^{(1)} + (1-\lambda)u^{(2)}) \leq \lambda L(u^{(1)}) + (1-\lambda)L(u^{(2)})$$

*Proof:* Define

$$(y^*, z^*) = \arg \max_{y \in \mathcal{Y}, z \in \mathcal{Z}} L(u^*, y, z)$$

where $u^* = \lambda u^{(1)} + (1-\lambda)u^{(2)}$. It follows that

$$L(u^*) = L(u^*, y^*, z^*)$$

In addition, note that

$$L(u^{(1)}, y^*, z^*) \leq \max_{y \in \mathcal{Y}, z \in \mathcal{Z}} L(u^{(1)}, y, z) = L(u^{(1)})$$

and similarly

$$L(u^{(2)}, y^*, z^*) \leq L(u^{(2)})$$

from which it follows that

$$\lambda L(u^{(1)}, y^*, z^*) + (1-\lambda)L(u^{(2)}, y^*, z^*) \leq \lambda L(u^{(1)}) + (1-\lambda)L(u^{(2)})$$

Finally, it is easy to show that

$$\lambda L(u^{(1)}, y^*, z^*) + (1-\lambda)L(u^{(2)}, y^*, z^*) = L(u^*, y^*, z^*) = L(u^*)$$

hence

$$L(u^*) \leq \lambda L(u^{(1)}) + (1-\lambda)L(u^{(2)})$$

which is the desired result. □

### A.2 Subgradients of $L(u)$

For any value of $u \in \mathbb{R}^d$, as before define

$$(y^{(u)}, z^{(u)}) = \operatorname*{argmax}_{y \in \mathcal{Y}, z \in \mathcal{Z}} L(u, y, z)$$

or equivalently,

$$y^{(u)} = \operatorname*{argmax}_{y \in \mathcal{Y}} \left( f(y) + \sum_{i,t} u(i,t) y(i,t) \right)$$

and

$$z^{(u)} = \operatorname*{argmax}_{z \in \mathcal{Z}} \left( g(z) - \sum_{i,t} u(i,t) z(i,t) \right)$$

Then if we define $\gamma^{(u)}$ as the vector with components

$$\gamma^{(u)}(i,t) = y^{(u)}(i,t) - z^{(u)}(i,t)$$

for $i \in \{1 \ldots n\}, t \in \mathcal{T}$, then $\gamma^{(u)}$ is a subgradient of $L(u)$ at $u$.

This result is a special case of the following theorem:[16]

**Theorem 10** *Define the function* $L : \mathbb{R}^d \to \mathbb{R}$ *as*

$$L(u) = \max_{i \in \{1 \ldots m\}} (a_i \cdot u + b_i)$$

*where* $a_i \in \mathbb{R}^d$ *and* $b_i \in \mathbb{R}$ *for* $i \in \{1 \ldots m\}$. *Then for any value of* $u$, *if*

$$j = \operatorname*{argmax}_{i \in \{1 \ldots m\}} (a_i \cdot u + b_i)$$

*then* $a_j$ *is a subgradient of* $L(u)$ *at* $u$.

*Proof:* For $a_j$ to be a subgradient at the point $u$, we need to show that for all $v \in \mathbb{R}^d$,

$$L(v) \geq L(u) + a_j \cdot (v - u)$$

Equivalently, we need to show that for all $v \in \mathbb{R}^d$,

$$\max_{i \in \{1 \ldots m\}} (a_i \cdot v + b_i) \geq \max_{i \in \{1 \ldots m\}} (a_i \cdot u + b_i) + a_j \cdot (v - u) \tag{39}$$

To show this, first note that

$$a_j \cdot u + b_j = \max_{i \in \{1 \ldots m\}} (a_i \cdot u + b_i)$$

hence

$$\max_{i \in \{1 \ldots m\}} (a_i \cdot u + b_i) + a_j \cdot (v - u) = b_j + a_j \cdot v \leq \max_{i \in \{1 \ldots m\}} (a_i \cdot v + b_i)$$

thus proving the theorem. □

### A.3 Convergence Proof for the Subgradient Method

Consider a convex function $L : \mathbb{R}^d \to \mathbb{R}$, which has a minimizer $u^*$ (i.e., $u^* = \operatorname{argmin}_{u \in \mathbb{R}^d} L(u)$). The subgradient method is an iterative method which initializes $u$ to some value $u^{(0)} \in \mathbb{R}^d$, then sets

$$u^{(k+1)} = u^{(k)} - \delta_k g_k$$

for $k = 0, 1, 2, \ldots$, where $\delta_k > 0$ is the stepsize at the $k$'th iteration, and $g_k$ is a subgradient at $u^{(k)}$: that is, for all $v \in \mathbb{R}^d$,

$$L(v) \geq L(u^{(k)}) + g_k \cdot (v - u^{(k)})$$

The following theorem will then be very useful in proving convergence of the method (the theorem and proof is taken from Boyd & Mutapcic, 2007):

16. To be specific, our definition of $L(u)$ can be written in the form $\max_{i \in \{1 \ldots m\}} (a_i \cdot u + b_i)$ as follows. Define the integer $m$ to be $|\mathcal{Y}| \times |\mathcal{Z}|$. Define $(y^{(i)}, z^{(i)})$ for $i \in \{1 \ldots m\}$ to be a list of all possible pairs $(y, z)$ such that $y \in \mathcal{Y}$ and $z \in \mathcal{Z}$. Define $b_i = f(y^{(i)}) + g(z^{(i)})$, and $a_i$ to be the vector with components $a_i(l, t) = y^{(i)}(l, t) - z^{(i)}(l, t)$ for $l \in \{1 \ldots n\}, t \in \mathcal{T}$. Then it can be verifed that $L(u) = \max_{i \in \{1 \ldots m\}} (a_i \cdot u + b_i)$.

**Theorem 11** *Assume that for all k,* $||g_k||^2 \leq G^2$ *where G is some constant. Then for any* $k \geq 0$*,*

$$\min_{i \in \{0 \ldots k\}} L(u^{(i)}) \leq L(u^*) + \frac{||u^{(0)} - u^*||^2 + G^2 \sum_{i=0}^{k} \delta_i^2}{2 \sum_{i=0}^{k} \delta_i}$$

*Proof:* First, given the updates $u^{(k+1)} = u^{(k)} - \delta_k g_k$, we have for all $k \geq 0$,

$$\begin{aligned} ||u^{(k+1)} - u^*||^2 &= ||u^{(k)} - \delta_k g_k - u^*||^2 \\ &= ||u^{(k)} - u^*||^2 - 2\delta_k g_k \cdot (u^{(k)} - u^*) + \delta_k^2 ||g_k||^2 \end{aligned}$$

By the subgradient property,

$$L(u^*) \geq L(u^{(k)}) + g_k \cdot (u^* - u^{(k)})$$

hence

$$-g_k \cdot (u^{(k)} - u^*) \leq L(u^*) - L(u^{(k)})$$

Using this inequality, together with $||g_k||^2 \leq G^2$, gives

$$||u^{(k+1)} - u^*||^2 \leq ||u^{(k)} - u^*||^2 + 2\delta_k \left( L(u^*) - L(u^{(k)}) \right) + \delta_k^2 G^2$$

Taking a sum over both sides of $i = 0 \ldots k$ gives

$$\sum_{i=0}^{k} ||u^{(i+1)} - u^*||^2 \leq \sum_{i=0}^{k} ||u^{(i)} - u^*||^2 + 2 \sum_{i=0}^{k} \delta_i \left( L(u^*) - L(u^{(i)}) \right) + \sum_{i=0}^{k} \delta_i^2 G^2$$

and hence

$$||u^{(k+1)} - u^*||^2 \leq ||u^{(0)} - u^*||^2 + 2 \sum_{i=0}^{k} \delta_i \left( L(u^*) - L(u^{(i)}) \right) + \sum_{i=0}^{k} \delta_i^2 G^2$$

Finally, using $||u^{(k+1)} - u^*||^2 \geq 0$ and

$$\sum_{i=0}^{k} \delta_i \left( L(u^*) - L(u^{(i)}) \right) \leq \left( \sum_{i=0}^{k} \delta_i \right) \left( L(u^*) - \min_{i \in \{0 \ldots k\}} L(u^{(i)}) \right)$$

gives

$$0 \leq ||u^{(0)} - u^*||^2 + 2 \left( \sum_{i=0}^{k} \delta_i \right) \left( L(u^*) - \min_{i \in \{0 \ldots k\}} L(u^{(i)}) \right) + \sum_{i=0}^{k} \delta_i^2 G^2$$

Rearranging terms gives the result in the theorem. □

This theorem has a number of consequences. As one example, for a constant step-size, $\delta_k = h$ for some $h > 0$,

$$\lim_{k \to \infty} \left( \frac{||u^{(0)} - u^*||^2 + G^2 \sum_{i=1}^{k} \delta_i^2}{2 \sum_{i=1}^{k} \delta_i} \right) = \frac{Gh}{2}$$

hence in the limit the value for

$$\min_{i \in \{1 \ldots k\}} L(u^{(i)})$$

is within $Gh/2$ of the optimal solution. A slightly more involved argument shows that under the assumptions that $\delta_k > 0$, $\lim_{k \to \infty} \delta_k = 0$, and $\sum_{k=0}^{\infty} \delta_k = \infty$,

$$\lim_{k \to \infty} \left( \frac{||u^{(0)} - u^*||^2 + G^2 \sum_{i=1}^{k} \delta_i^2}{2 \sum_{i=1}^{k} \delta_i} \right) = 0$$

See Boyd and Mutapcic for the full derivation.

### A.4 Proof of Theorem 6

Recall that our goal is to prove that

$$\max_{y \in \mathcal{Y}} f(y) = \max_{\alpha \in \Delta_y} \sum_{y \in \mathcal{Y}} \alpha_y f(y)$$

We will show this by proving: (1) $\max_{y \in \mathcal{Y}} f(y) \leq \max_{\alpha \in \Delta_y} \sum_{y \in \mathcal{Y}} \alpha_y f(y)$, and (2) $\max_{y \in \mathcal{Y}} f(y) \geq \max_{\alpha \in \Delta_y} \sum_{y \in \mathcal{Y}} \alpha_y f(y)$.

First, consider case (1). Define $y^*$ to be a member of $\mathcal{Y}$ such that

$$f(y^*) = \max_{y \in \mathcal{Y}} f(y)$$

Next, define $\alpha_{y^*} = 1$, and $\alpha_y = 0$ for $y \neq y^*$. Then we have

$$\sum_{y \in \mathcal{Y}} \alpha_y f(y) = f(y^*)$$

Hence we have found a setting for the $\alpha$ variables such that

$$\sum_{y \in \mathcal{Y}} \alpha_y f(y) = \max_{y \in \mathcal{Y}} f(y)$$

from which it follows that

$$\max_{\alpha \in \Delta_y} \sum_{y \in \mathcal{Y}} \alpha_y f(y) \geq \max_{y \in \mathcal{Y}} f(y)$$

Next, consider case (2). Define $\alpha^*$ to be a setting of the $\alpha$ variables such that

$$\sum_{y} \alpha^*_y f(y) = \max_{\alpha \in \Delta_y} \sum_{y \in \mathcal{Y}} \alpha_y f(y)$$

Next, define $y^*$ to be a member of $\mathcal{Y}$ such that

$$f(y^*) = \max_{y : \alpha^*_y > 0} f(y)$$

It is easily verified that

$$f(y^*) \geq \sum_{y} \alpha^*_y f(y)$$

Hence we have found a $y^* \in \mathcal{Y}$ such that

$$f(y^*) \geq \max_{\alpha \in \Delta_y} \sum_{y} \alpha^*_y f(y)$$

from which it follows that

$$\max_{y \in \mathcal{Y}} f(y) \geq \max_{\alpha \in \Delta_y} \sum_{y} \alpha^*_y f(y)$$

□